\documentclass[iicol, pdflatex,sn-mathphys-num]{sn-jnl}

\usepackage{graphicx}%
\usepackage{multirow}%
\usepackage{amsmath,amssymb,amsfonts}%
\usepackage{amsthm}%
\usepackage{mathrsfs}%
\usepackage[title]{appendix}%
\usepackage{xcolor}%
\usepackage{textcomp}%
\usepackage{manyfoot}%
\usepackage{booktabs}%
\usepackage{algorithm}%
\usepackage{algorithmicx}%
\usepackage{algpseudocode}%
\usepackage{listings}%
\usepackage{adjustbox}
\usepackage{tikz}
\usepackage{mathabx}
\usepackage{anyfontsize}
\usetikzlibrary{spy}

\newcommand{\Rb}{\ensuremath{\mathbb{R}}}
\def\prox{\text{prox}}
\def\div{\text{div}}

\DeclareMathOperator*{\argmin}{arg\,min}

\newlength{\spyimagewidth}
\newlength{\spyimageheight}
\newcommand{\spyimage}[2][0.20\linewidth]{%
  \setlength{\spyimagewidth}{#1}
  \setlength{\spyimageheight}{\spyimagewidth}
  \begin{tikzpicture}[spy using outlines={rectangle, magnification=1.8, size=0.4\spyimagewidth, connect spies,color=white}]
    \node[inner sep=0pt, anchor=south west] (image) at (0,0) {\adjustbox{trim={.05\width} {0.2\height} {0.45\width} {0.2\height},clip,width=\spyimagewidth}{\includegraphics{#2}}};
    \coordinate (spy point) at (0.85\spyimagewidth , 0.63\spyimagewidth );
    \coordinate (spy node) at (0.1\spyimagewidth, 0.8\spyimageheight);
    \spy on (spy point) in node at (spy node);
  \end{tikzpicture}%
}

\newcommand{\spyimagelol}[2][0.20\linewidth]{%
  \setlength{\spyimagewidth}{#1}
  \setlength{\spyimageheight}{\spyimagewidth}
  \begin{tikzpicture}[spy using outlines={rectangle, magnification=1.8, size=0.4\spyimagewidth, connect spies,color=white}]
    \node[inner sep=0pt, anchor=south west] (image) at (0,0) {\adjustbox{trim={.1\width} {0.1\height} {0.1\width} {0.1\height},clip,width=\spyimagewidth}{\includegraphics{#2}}};
    \coordinate (spy point) at (0.45\spyimagewidth , 0.43\spyimagewidth );
    \coordinate (spy node) at (0.1\spyimagewidth, 0.8\spyimageheight);
    \spy on (spy point) in node at (spy node);
  \end{tikzpicture}%
}

\newcommand{\spyimagelamp}[2][0.20\linewidth]{%
  \setlength{\spyimagewidth}{#1}
  \setlength{\spyimageheight}{\spyimagewidth}
  \begin{tikzpicture}[spy using outlines={rectangle, magnification=2.7, size=0.4\spyimagewidth, connect spies,color=white}]
    \node[inner sep=0pt, anchor=south west] (image) at (0,0) {\adjustbox{trim={.1\width} {0.1\height} {0.1\width} {0.1\height},clip,width=\spyimagewidth}{\includegraphics{#2}}};
    \coordinate (spy point) at (0.62\spyimagewidth , 0.65\spyimagewidth );
       \coordinate (spy node) at (0.1\spyimagewidth, 0.85\spyimageheight);
    \spy on (spy point) in node at (spy node);
  \end{tikzpicture}%
}

\begin{document}

\title[Article Title]{Nonlocal Retinex-Based Variational Model and its Deep Unfolding Twin for Low-Light Image Enhancement }


\author*[1]{\fnm{Daniel} \sur{Torres}}\email{daniel.torres@uib.es}

\author[1]{\fnm{Joan} \sur{Duran}}\email{joan.duran@uib.es}

\author[1]{\fnm{Julia} \sur{Navarro}}\email{julia.navarro@uib.es}

\author[1]{\fnm{Catalina} \sur{Sbert}}\email{catalina.sbert@uib.es}

\affil*[1]{\orgdiv{Dpt.~Mathematics and Computer Science and IAC3}, \orgname{Universitat de les Illes Balears}, \orgaddress{\street{Cra.~de Valldemossa, km.~7.5}, \city{Palma}, \postcode{07122}, \state{Illes Balears}, \country{Spain}}}


\abstract{Images captured under low-light conditions present significant limitations in many applications, as poor lighting can obscure details, reduce contrast, and hide noise. Removing the illumination effects and enhancing the quality of such images is crucial for many tasks, such as image segmentation and object detection. In this paper, we propose a variational method for low-light image enhancement based on the Retinex decomposition into illumination, reflectance, and noise components. A color correction pre-processing step is applied to the low-light image, which is then used as the observed input in the decomposition. Moreover, our model integrates a novel nonlocal gradient-type fidelity term designed to preserve structural details. Additionally, we propose an automatic gamma correction module. Building on the proposed variational approach, we extend the model by introducing its deep unfolding counterpart, in which the proximal operators are replaced with learnable networks. We propose cross-attention mechanisms to capture long-range dependencies in both the nonlocal prior of the reflectance and the nonlocal gradient-based constraint. Experimental results demonstrate that both methods compare favorably with several recent and state-of-the-art techniques across different datasets. In particular, despite not relying on learning strategies, the variational model outperforms most deep learning approaches both visually and in terms of quality metrics.
}




\maketitle

\section{Introduction}\label{sec1}

Low-light image enhancement \cite{liu2021benchmarking, anoop2024advancements} has received significant attention in computer vision, since poor image visibility can negatively impact the performance of various applications. This task remains challenging, as it requires the simultaneous adjustment of color, contrast, brightness, and noise. While advanced photographic techniques and professional equipment can improve visual quality, they cannot fully prevent the amplification of noise hidden in dark regions, and some details may still be buried in the darkness. To address these challenges, researchers have worked on developing algorithms from multiple perspectives.

Classical approaches encompass a wide range of strategies, including histogram equalization \cite{He1, PAUL2022}, Retinex-based methods \cite{kimmel, Ng, LIME} , multi-exposure fusion \cite{12, Buades_lisani_petro_sbert}, and defogging model techniques \cite{14,15}. Among these, Retinex-based models are particularly predominant. The Retinex theory \cite{Retinex} explains how the human visual system perceives color independently of global illumination changes. This is typically modeled  by assuming that an image is the product of the illumination and the reflectance.

Decomposing an image into  the reflectance and illumination components is an ill-posed inverse problem that requires prior knowledge. In the variational setting, the solution is obtained by minimizing an energy functional that comprises data-fidelity and regularization terms. Several authors \cite{kimmel, LIME} have proposed variational models to only compute the illumination, while others aim to recover both components simultaneously \cite{Ng, structure-Retinex}.

The effectiveness of variational techniques depends on designing appropriate priors, a task that has proven challenging over time. Early methods assumed simple regularization terms to enforce smooth illumination and piecewise-constant reflectance, typically penalizing gradient oscillations using $L^2$ and $L^1$ norms, respectively \cite{13}. Afterwards, more sophisticated regularizers have been proposed to improve the decomposition, including low-rank \cite{LR3M}, nonlocal \cite{torres2024combining}, and fractional gradient \cite{fractional2023} priors.

The rise of deep learning has led to an increasing number of enhancement methods, which can be categorized into purely deep learning-based techniques \cite{mbllen, ZeroDCE, RetinexFormer} and model-based deep unfolding approaches \cite{RUAS, URetinexNet, RAUNA}. The first category directly employs neural networks to learn natural priors, relying on complex architectures and being highly dependent on training data. These limitations have motivated the development of a new class of networks that integrate model-based energy formulations into deep learning frameworks, resulting in more efficient and interpretable architectures.

In this paper, we propose a low-light image enhancement variational model based on the Retinex decomposition into luminance, reflectance, and noise components. The noise component plays a crucial role in preventing the amplification of noise during gamma correction and enhancement.  We impose nonlocal total variation sparsity on the reflectance and total variation sparsity on the illumination. To mitigate color degradation, we introduce a color correction pre-processing step for the low-light image, which is then used as the observed input in the decomposition model. Furthermore, low-light images suffer from reduced contrast, which is typically associated with small gradients. To address this issue, we introduce a nonlocal gradient-type constraint that enforces similarity between the gradient of the reflectance and an adjusted version of the input image. The similarity weights are specifically designed to capture the structural details within the image gradient. Additionally, we propose an automatic gamma correction module, which reduces the model's parameter load. Since the noise is explicitly addressed within the decomposition, no post-processing is required. 

We then extend the proposed variational model by introducing its deep unfolding counterpart, learning priors for the reflectance and illumination components, as well as the weights of the nonlocal gradient-type constraint. Inspired by SWIN transformers \cite{swir-transformer}, we introduce cross-attention mechanisms by modifying multi-head attention \cite{multihead_ivan} to capture long-range dependencies adjusting the keys and queries in each head, mimicking the behaviour of nonlocal operators. The proposed unfolding approach does not require pre-processing or gamma correction modules. 

To summarize, the main contributions of this paper are the following:
\begin{itemize}
    \item A pre-processing color correction step for the low-light image to mitigate color distortions. The corrected image is then used in the decomposition model that separates images into luminance, reflectance, and noise components.
    \item A variational model based on the previous decomposition, which imposes a nonlocal total variation prior on the reflectance and total variation on the luminance, while introducing a nonlocal gradient-type constraint to enhance contrast. We also adapt the first-order primal-dual algorithm by Chambolle and Pock \cite{Chambolle} to the resulting nonconvex energy, enabling the computation of a local minimizer.
    \item An automatic gamma correction module.
    \item A deep unfolding framework that learns the priors for the reflectance and illumination components, and proposes cross-attention mechanisms to emulate the behaviour of nonlocal operators. This approach eliminates the need for pre-processing the low-light image or applying gamma correction to the final output.
\end{itemize}

This work extends our previous conference paper \cite{icip_dani}, where classical Tikhonov regularization was applied to the illumination, total variation was used for the reflectance, and the nonlocal-gradient constraint was introduced. The main differences in this paper include the introduction of the pre-processing color correction step, the design of a deep unfolding counterpart, and the comprehensive benchmarking and ablation study.

The rest of the paper is organized as follows. In Section 2, we review the state of the art in low-light image enhancement. Section 3 introduces the proposed variational method, while in Section 4 we present its unfolded counterpart. An exhaustive performance comparison across LOLv2, LOLv2-Synthetic and LIME datasets is presented in Section 5. Section 6 conducts an ablation study that highlights the importance of the selected variational terms and network architecture. Finally, conclusions are drawn in Section 7.

\section{State of the Art}

\subsection{Classical methods}

The earliest attempts to enhance low-light images focused on manipulating pixel intensity by directly transforming its grayscale value, either through a linear adjustment or by applying nonlinear functions such as logarithmic or gamma corrections. However, these techniques do not consider the overall gray-level distribution of the image. As a result, histogram equalization strategies \cite{He1, PAUL2022} were developed to adjust the gray values of single pixels using the cumulative distribution function.

Other methods are based on image fusion. The difficulty in obtaining images of a scene over time or under different lighting conditions is why the most popular approaches propose fusing different estimations based on a single image. In this setting, Fu et al.~\cite{12} combine several illumination maps generated by different enhancement techniques. Similarly, Buades et al.~\cite{Buades_lisani_petro_sbert} use several tone mappings and estimate the enhanced image through a multiscale fusion strategy. 

Retinex theory \cite{Retinex} simulates how the human visual system perceives color independently of global illumination changes. One of the most widely used models assumes that the observed image $I$ is the product of the illumination $L$, which depicts the light intensity on the objects, and the reflectance $R$, which represents their physical characteristics:
\begin{equation}\label{RL}
I=R\circ L,
\end{equation}
where $\circ$ denotes pixel-wise product.  Recovering $L$ and $R$ from \eqref{RL} is an ill-posed inverse problem that requires prior knowledge. Typically, $L$ is assumed to be smooth, while $R$ contains fine details and texture. 

Different approaches have been proposed within the Retinex framework, such as patch-based \cite{Retinex}, partial differential equations \cite{Morel}, and center/surround \cite{9} methods. In particular, single-scale Retinex \cite{9} and multiscale Retinex \cite{10} may be considered as seminal works. These methods filter the input image using Gaussian kernels, taking the low-frequency result as the illumination component and the residual image as the reflectance component. 

In \cite{dehaze-retinex}, the authors show a duality between the Retinex theory and dehazing techniques. They prove that Retinex on inverted intensities solves the image dehazing problem. Based on this duality, dehazing models \cite{14,15} produce competitive results for the enhancement of a low-light image.
 
However, the most common approach to tackle ill-posed inverse problems like \eqref{RL} is to use variational techniques, which assume a certain regularity in the image. The solution is obtained by minimizing an energy functional that consists of regularization and fidelity terms.

Kimmel et al.~\cite{kimmel} pioneered a variational model to compute the illumination, enforcing spatial smoothness. Ng et al.~\cite{Ng} proposed estimating illumination and reflectance simultaneously, using the total variation (TV) to directly compute the reflectance. These methods linearize equation \eqref{RL} by applying logarithms, but errors in gradient-type energy terms are amplified. In \cite{13}, weights are introduced to address issues arising from large gradients when either $R$ or $L$ is small.

One of the most celebrated works is LIME \cite{LIME}. Guo et al. proposed estimating the illumination at each pixel as the maximum value per channel and refining it through the minimization of a simple TV-based energy. Then, the reflectance is computed directly using equation \eqref{RL}. A post-processing step is finally applied to reduce noise.

Li et al.~\cite{structure-Retinex} introduced the noise component in the decomposition:
\begin{equation}\label{RL+N}
I=R\circ L + N,
\end{equation}
which helps prevent noise amplification during enhancement. The authors also adopted a fidelity term for the gradient of the reflectance to preserve structural details. Ren et al.~\cite{LR3M} additionally minimized the rank of matrices representing similar patches in the reflectance. They omitted the noise component, assuming instead that it is part of the reflectance. In \cite{fractional2023}, Chen et al. followed the previous approach incorporating fractional-order priors to obtain various gradients flexibly.

Nonlocal regularization \cite{nlosher,duran2014nonlocal} allows any point to interact directly with any other point in the whole domain and computes the distance between them in terms of closeness of intensity values in the image. Therefore, the underlying assumption behind nonlocal regularization is that images are self-similar, making it a good prior to preserve structure and details. In this setting, Zosso et al.~\cite{nonlocalretinex} combined earlier Retinex models with nonlocal regularization. Nevertheless, the continuation of these techniques in recent research has been rather limited \cite{torres2024combining}.

\subsection{Purely deep learning methods}

Purely deep learning strategies are distinguished by their specific architecture, which plays a crucial role in the performance of the method. Early approaches focused on learning the complex relationship between low-light and enhanced images \cite{LLNet}. 
In \cite{mbllen}, Lv et al. designed a multi-branch network where the features are enhanced and fused. The same authors \cite{attention-guided} improved the architecture by introducing an attention map and a noise map into the feature enhancement module. Jiang et al.~\cite{enlightengan} employed generative adversarial learning, where discriminators are constructed to directly map a low-light image to a normal-light image.

Retinex-based deep learning methods have attracted much more attention  due to their explicit physical meaning. Wei et al.~\cite{deep-retinex} introduced CNNs to adjust each component of the Retinex decomposition separately. However, reflectance restoration is treated using a classical denoising algorithm. Zhang et al. \cite{KinDplus} constructed a network divided into three modules: layer decomposition, reflectance restoration, and illumination adjustment. In \cite{sparse}, a sparse gradient regularization is incorporated at the decomposition stage. Cai et al.~\cite{RetinexFormer} introduced a one-stage Retinex-based framework that simplifies the enhancement process by estimating illumination to brighten the image and then restoring any corruption with an illumination-guided transformer. 

In contrast to networks that learn only single illumination or mapping relationships, Ghillie \cite{Ghillie} proposed a multi-illumination estimation framework based on ghost imaging theory, along with denoising and color restoration networks.

To solve the problem of the availability of training data for Retinex decomposition, several unsupervised networks have been proposed. Guo et al.~\cite{ZeroDCE} introduced an end-to-end network producing high-order curves used for pixel-wise adjustment.
In \cite{UnsupervisedNightImageEnhancement}, the authors integrated a network to decompose shading, reflectance and light-effects layers guided by prior losses.

\subsection{Model-based deep unfolding methods}

Data-driven approaches can learn natural priors but their effectiveness depends on complex architectures, making the networks less flexible and harder to interpret. Moreover, they cannot benefit from the physic-based constraints imposed in variational models. To leverage the strengths of both, explainable networks have been designed by unrolling the optimization scheme derived from minimizing a Retinex-based energy into a deep learning framework \cite{RUAS, RIRO, RAUNA}. By focusing on modeling specific operations instead of the entire problem, these methods commonly offer simpler and interpretable architectures.

Liu et al.~\cite{RUAS} introduced the unrolling methodology in the context of low-light image enhancement. However, their approach relied on the network architecture search for the design of the network structure, and they ignored the interaction between illumination and reflectance, which is essential for an accurate decomposition.

In \cite{URetinexNet}, Wu et al.~unfolded the Retinex decomposition model
\begin{equation}\label{uretinexnet_model}
    \min_{R,L} \|I-R\circ L\|_2^2+\alpha\Phi(R)+\beta\Psi(L)
\end{equation}
to integrate physical priors of $R$ and $L$ into the network structure. First, an initialization module is designed to improve the effectiveness of the unfolding optimization scheme and generate clear reflectance using the normal-light image. Then, they unroll the iterative algorithm for solving the minimization problem and implicitly embed each of $\Phi$ and $\Psi$ into a network module. Finally, they include an illumination adjustment module incorporating a gradient fidelity term in the loss function, but the light enhancement parameter must be specified by the user. An improved version of the architecture was proposed in \cite{uretinex++} by introducing a cross-stage fusion block to correct color defects and a spatial consistency loss function for the illumination adjustment module.

Liu et al.~\cite{RAUNA} based their unfolding approach on the model \eqref{uretinexnet_model}, but they introduced an additional gradient fidelity term as follows:
\begin{equation*}
    \min_{R,L} \frac{1}{2}\|I-R\circ L\|_2^2+\alpha\Phi(R)+\beta\Psi(L)+\frac{\mu}{4}\|\nabla R-G\|_{2}^2,
\end{equation*}
where $G$ is the amplified gradient of $I$ introduced in  \cite{structure-Retinex}.
After the Retinex decomposition using the unfolding method, they incorporate an illumination module to adjust the illumination. However, it depends on a global brightness parameter to be specified by the user in the absence of ground truth. To solve this, they propose a self-supervised strategy to fine-tune the adjustment networks at test time.

Zhao et al.~\cite {RIRO} presented a new deep unfolding network, in which the first part consists of low-light decomposition and enhancement modules with the goal of obtaining clear illumination and reflectance components. Then, they formulated the image reconstruction problem as 
\begin{equation*}
\begin{aligned}
    \min_{R,L} \:&\|R-R_{\text{low}}\|_2^2 + \|L-L_{\text{high}}\|_2^2+\|R\circ L-I\|_2^2 \\&+\alpha\Phi(R)+\beta\Psi(L).
    \end{aligned}
\end{equation*}
In the unfolding step, the solutions are obtained using sub-networks with a dual-domain proximal block instead of classical residual networks.

Many of these unfolding methods \cite{uretinex++, RAUNA, RIRO} employ an initial decomposition module to process both the low-light and reference images. This preliminary step plays a crucial role in establishing a well-defined decomposition that enhances the effectiveness of the unfolding process and in providing reference illumination and reflectance to guide the iterative optimization.

\section{Proposed Variational Method}

We propose a variational model built on the decomposition given in \eqref{RL+N}. The noise component plays a crucial role in preventing the amplification of noise during gamma correction and enhancement. Furthermore, low-light images suffer from reduced contrast, which is typically associated with small gradients. To address this issue, we introduce a nonlocal energy term that enforces similarity between the gradient of the reflectance and an adjusted version of the input image. 

\subsection{Definitions and notations}

Let us denote the low-light image as $I\in\Rb^{C\times M}$, where $M$ is the number of pixels and $C$ is  the number of color channels, and the reflectance, noise, and illumination components as $R, N\in\Rb^{C\times M}$ and $L\in\Rb^{M}$, respectively. The corresponding gradients $\nabla R, \nabla I \in\Rb^{C\times M\times 2}$ and $\nabla L \in\Rb^{M\times 2}$ are computed via forward differences with Neumann boundary conditions. We use the indices $i,j\in \{1,\dots, M\}$ for pixels, $k\in \{1,\dots, C\}$ for channels, and $t\in\{1,2\}$ for gradient components. For example, $(\nabla R)_{k,i,t}$ denotes the $t$-th component of the gradient of the $k$-th channel of the reflectance at pixel $i$.
We also consider $\Rb^{n_i\times n_j\times n_k\times n_t}$ endowed with the norm $\|x\|_{s_t,s_k,s_j,s_i}$, which is defined as
\begin{equation*}
\left(\hspace{-0.07cm}\sum_{i=1}^{n_i}\hspace{-0.07cm}\left(\hspace{-0.07cm}\sum_{j=1}^{n_j}\hspace{-0.07cm}\left(\hspace{-0.05cm}\sum_{k=1}^{n_k}\hspace{-0.08cm}\left(\sum_{t=1}^{n_t}\left|x_{i,j,k,t}\right|^{s_t}\hspace{-0.03cm}\right)^{s_k/s_t}\hspace{-0.03cm}\right)^{s_j/s_k}\hspace{-0.03cm}\right)^{s_i/s_j}\hspace{-0.03cm}\right)^{1/s_i}\hspace{-0.15cm}.
\end{equation*}
If $s_i=s_j=s_k=s_t$, we denote the norm simply by $\|x\|_{s}$. Normed spaces of lower dimensions are defined analogously.

Let $\omega\in \Rb^{M\times M}$ be a non-negative weight function, assumed to be the same for all channels. The nonlocal gradient $\nabla_{\omega}{R}\in\Rb^{C\times M\times M}$ is defined for each channel $k$ and each pair of pixels $i$ and $j$ as
$$
\left(\nabla_{\omega} R\right)_{k,i,j}=
\sqrt{\omega_{i,j}}\left(R_{k,i}-R_{k,j}\right).
$$
The associated nonlocal divergence $\text{\div}_{\omega}{p}\in\Rb^{C\times M}$ of a variable $p\in\Rb^{C\times M\times M}$ is thus given  by
\begin{equation}\label{eq-div-nltv}
        \left(\text{div}_{\omega}{p}\right)_{k,i}
        =\sum_{j=1}^M \left(p_{k,i,j}\sqrt{\omega_{i,j}}-p_{k,j,i}\sqrt{\omega_{j,i}}\,\right).
\end{equation}

\subsection{Nonlocal Retinex-based variational model}

We propose to simultaneously estimate the illumination, reflectance, and noise by solving the following minimization problem:
\begin{equation}\label{energia1}
\begin{aligned}
&\min_{R,L,N} \frac{1}{2}\|R\circ L+N-\tilde{I}\|_{2}^2+\alpha\|\nabla_\omega R\|_{2,1,2}\\&+\frac{\beta}{2} \|\nabla L\|_{2,1}  +\frac{\lambda}{2} \|N\|_{2}^2 +\frac{\mu}{2} \|(\nabla R-\nabla \hat{I})_{\hat{\omega}}\|_{2}^2,
\end{aligned}
\end{equation}
where $\alpha,\beta, \lambda, \mu>0$ are trade-off parameters. The first term corresponds to the decomposition model but considers $\tilde{I}\in\Rb^{C\times M}$, a color corrected version of the low-light image computed following the procedure described in Subsection \ref{ColorCorrection}, instead of $I$. The $\alpha$-term enforces nonlocal total variation sparsity, which serves as a useful prior for preserving fine details and texture in the reflectance component, the $\beta$-term promotes total variation in the illumination component to reduce noise and reconstruct the main geometrical structure, while the $\lambda$-term constrains the amount of noise.

The $\mu$-term is a newly proposed nonlocal gradient-type constraint that minimizes the nonlocal distance between the gradient of the reflectance and the gradient of $\hat{I}\in\Rb^{C\times M}$, a pre-processed version of $\tilde{I}$. This pre-processing involves applying BM3D to $\tilde{I}$ for denoising \cite{dabov2007image}, followed by gamma correction on each channel, as described in Subsection \ref{GammaIdeal}. In this setting, $\hat{\omega}\in\Rb^{M\times M\times 2}$ is a non-negative weight function, assumed to be the same across channels. The nonlocal vector $(\nabla R-\nabla \hat{I})_{\hat{\omega}}\in \Rb^{C\times M\times M\times 2}$ is defined for each channel $k$, each component $t$ and at each pair of pixels $i$ and $j$ as 
$$
\big((\nabla R-\nabla \hat{I})_{\hat{\omega}}\big)_{k,i,j,t}=\sqrt{\hat{\omega}_{i,j,t}}((\nabla R)_{k,i,t}-(\nabla\hat{I})_{k,j,t}).
$$

\subsection{Color correction}\label{ColorCorrection}

Low-light images suffer from color degradation. Since color restoration primarily relies on the Retinex decomposition model, this issue will persist unless addressed. Therefore, we propose introducing a color-corrected version of the low-light image $I$. 

In \cite{ancuti}, the authors compensate for the predominance of the green channel in underwater images by adding a fraction of this channel to the red one. Following this idea, we propose a proportional compensation for channels whose mean value deviates most from $0.5$. Let us denote the mean value of $I$ for each channel $k$ as $M_k$, and define $n_{\text{min}}=\argmin_{k} \{|M_k-0.5|\}$. The color correction is applied to each channel $k\neq n_{\text{min}}$ as
$$\tilde{I}_k=I_k+\vartheta(M_{n_{\text{min}}}-M_k)(1-I_k)I_{n_{\text{min}}},$$
where $\vartheta$ is a proportional factor.

\begin{figure*}[t]
\begin{minipage}[b]{\linewidth}
  \centering
  {\includegraphics[width=\textwidth]{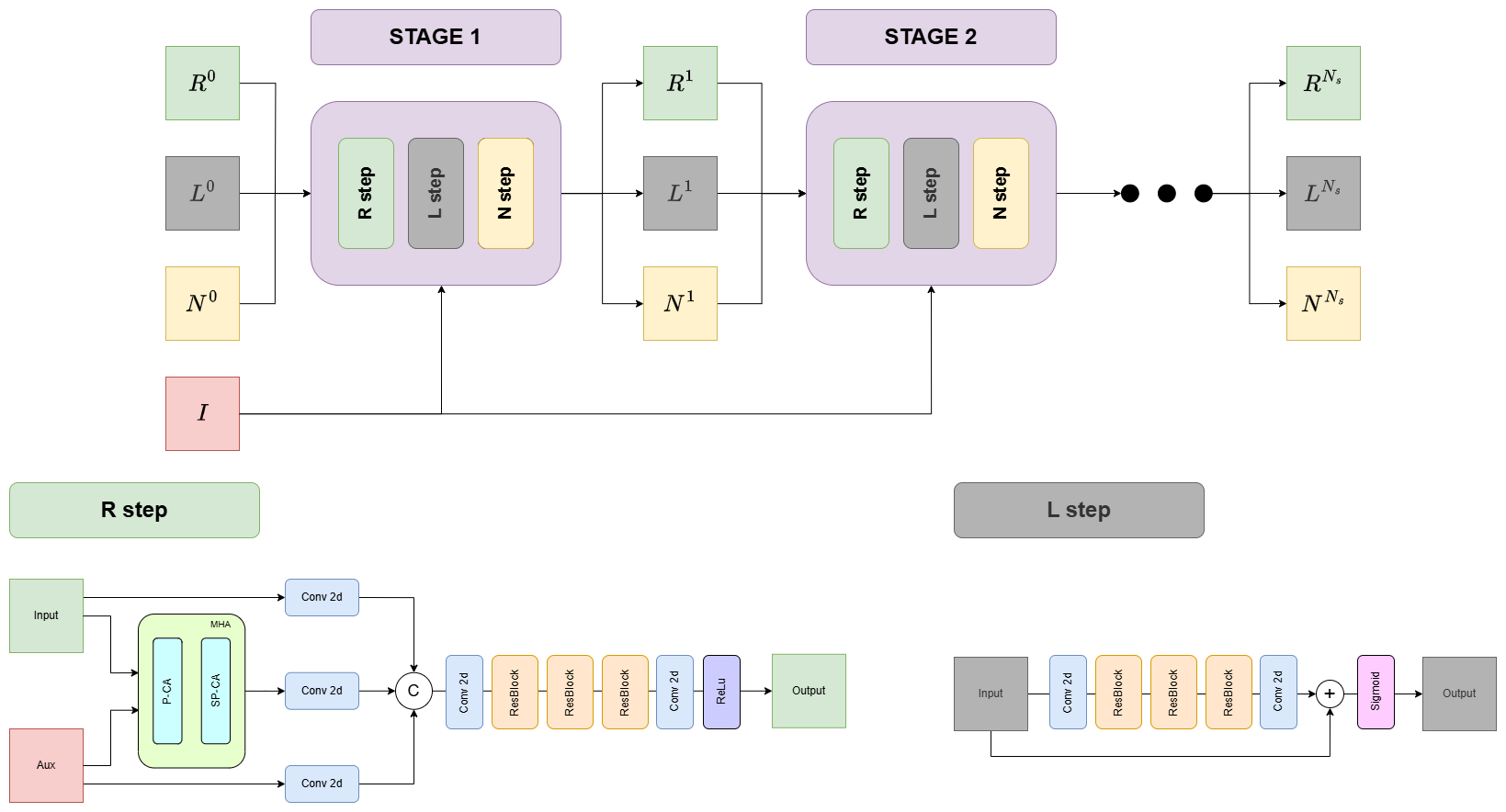}}
\end{minipage}
\hfill
\caption{Overall procedure of the proposed unfolding method. Each ResBlock consists of two convolutional layers followed with a residual connection.}
\label{fig:unfolding}
\end{figure*}

\subsection{Nonlocal weights}

For the weights $\omega\in \Rb^{M\times M}$ involved in the nonlocal total variation of the reflectance, we propose to consider both the spatial closeness between points and the similarity in the color-corrected image $\tilde{I}$. To gain robustness in the comparison, the similarity distance is evaluated by considering a whole patch around each pixel. Additionally, for computational efficiency, nonlocal interactions are limited to pixels within a certain distance. In practice, the weights are defined as
\begin{equation}\label{classical_weights}
\omega_{i,j}=\frac{1}{\Gamma_i}\exp\left(-\dfrac{|i-j|^2}{h^2_{\text{spt}}}-\dfrac{d(\tilde{I}_{:,i},\tilde{I}_{:,j})}{h^2_{\text{sim}}}\right)
\end{equation}
if $j\in B(i,\nu)\cap\mathbb{Z}$, and zero otherwise. In this setting, $\nu\in\mathbb{Z}^+$ determines the size of the search window, $h_{\text{spt}}, h_{\text{sim}}>0$ are filtering parameters that control how fast the weights decay with increasing spatial distance or dissimilarity between patches, respectively, $\Gamma_i$ is the normalizing factor, and
\begin{equation*}
d\big(\tilde{I}_{:,i}, \tilde{I}_{:,j}\big)=\hspace{-0.1cm}\sum_{z\in B(0, \kappa)\cap\mathbb{Z}}  \hspace{-0.1cm}|\tilde{I}_{:,i+z}-\tilde{I}_{:,j+z}|^2
\end{equation*} 
is the squared Euclidean distance between color patches of size $(2\kappa +1)\times (2\kappa+1)$ centered at pixels $i$ and $j$. In the end, the dimension of the nonlocal gradient reduces to $\nabla_{\omega} R\in\Rb^{C\times M\times (2\nu+1)^2}$.
Finally, to prevent excessive influence of the reference pixel, $\omega_{i,i}$ is set to the maximum of the weights within the search window for $j\neq i$.

The classical nonlocal weights described before quantify the similarity between any pair of pixels. However, since our objective with the nonlocal gradient-type fidelity term is to strengthen the structural information hidden in the low-light image, we argue that relying on pixel intensities is not the most effective approach. Instead, we propose searching within its image gradient. However, since gradient computation is highly sensitive to noise, which can be amplified in dark images, we propose searching for similarities in $\nabla \hat{I}$.

Therefore, the weights $\hat{\omega}\in\Rb^{M\times M\times 2}$ are defined for each direction $t\in\{1,2\}$ as
\begin{equation*}
\hat{\omega}_{i,j,t}=\frac{1}{\Gamma_{i,t}}\exp\left(-\dfrac{d((\nabla \hat{I})_{:,i,t},(\nabla \hat{I})_{:,j,t})}{\hat{h}^2_{\text{sim}}}\right)
\end{equation*}
if $j\in B(i,\hat{\nu})\cap\mathbb{Z}$, and zero otherwise. Now, $\hat{\nu}\in\mathbb{Z}^+$ determines the size of the search window, $\hat{h}_{\text{sim}}>0$ measures how fast the weights decay with increasing dissimilarity between patches, $\hat{\Gamma}_{i,t}$ is the normalizing factor, and
\begin{equation*}
d\big((\nabla \hat{I})_{:,i,t},(\nabla \hat{I})_{:,j,t}\big)=\hspace{-0.5cm}\sum_{z\in B(0, \hat{\kappa})\cap\mathbb{Z}} \hspace{-0.15cm} |(\nabla \hat{I})_{:,i+z,t}-(\nabla \hat{I})_{:,j+z,t}|^2
\end{equation*} 
is the squared Euclidian distance between color patches of size $(2\hat{\kappa} +1)\times (2\hat{\kappa}+1)$ centered at pixels $i$ and $j$. Again, $\hat{\omega}_{i,i,t}$ is set to the maximum of the weights within the search window for $i\neq j$. After all, the dimension of the nonlocal gradient-type vector in the $\mu$-term simplifies to $(\nabla R-\nabla \hat{I})_{\hat{\omega}}\in \Rb^{C\times M\times (2\hat{\nu}+1)^2\times 2}$.

\subsection{Saddle-point formulation and optimization}

The first-order primal-dual algorithm by Chambolle and Pock \cite{Chambolle} computes the minimizer of (possibly non-smooth) convex energies by reformulating the problem as a saddle-point optimization using dual variables. For this, one commonly relies on the fact that the convex conjugate of a norm is the indicator function of the unit dual norm ball and that any proper, convex, and lower-semicontinuous function is equal to its second convex conjugate \cite{chambolle2016introduction}. Since the $\alpha$-, $\beta$-, and $\mu$-terms in \eqref{energia1} are convex, we can dualize them and rewrite the problem as
$$
\begin{aligned}
&\min_{L,R,N}  \max_{p, q, o} \frac{1}{2}\|R\circ L+N-\tilde{I}\|^2_{2} + \langle \nabla_\omega R, p\rangle -\delta_{\mathcal{P}}(p) \\&+\langle \nabla L, o\rangle
 -\delta_{\mathcal{O}}(o)   +\langle (\nabla R-\nabla \hat{I})_{\hat\omega}, q\rangle  -\frac{1}{2\mu }\|q\|^2_{2}\\&+ \lambda \|N\|^2_{2},
\end{aligned}
$$
with $\mathcal{P}=\{p\in\Rb^{C\times M\times (2\nu+1)^2} : \|p\|_{2,\infty, 2}\leq \alpha\}, \mathcal{O}=\{o\in\Rb^{M\times 2} : \|o\|_{2, \infty}\leq \beta\}$, $q\in \Rb^{C\times M\times (2\hat{\nu}+1)^2\times 2}$, and $\delta$ is the indicator function.

\begin{table}[t]
\centering
\begin{tabular}{lccc}
  Method   & LPIPS $\downarrow$ & SSIM $\uparrow$ & PSNR $\uparrow$\\ \hline
        Pure deep learning \\ \hline
   RetinexNet \cite{deep-retinex}  & 0.474 & 0.7517  & 15.80 \\
       ZeroDCE \cite{ZeroDCE} & 0.335 & 0.7714 & 13.33 \\
AGLLNet \cite{attention-guided} & 0.210 & 0.8400  & 17.44 \\
EnlightenGAN \cite{enlightengan}  & 0.322 &  0.829 & 17.32 \\ 
        KinD+ \cite{KinDplus}  & 0.187 & 0.8553 & 16.64 \\ 
                Night-Enhancement \cite{UnsupervisedNightImageEnhancement} & 0.241 & 0.8736  & 21.08 \\
        Retinexformer \cite{RetinexFormer}  & 0.147 & \underline{0.9234}  & \underline{22.45} \\
        Ghillie \cite{Ghillie}  & 0.193 & 0.8886  & 19.69 \\
        \hline 
        Unfolding\\ \hline
          RUAS \cite{RUAS} & 0.270 & 0.7644  & 16.32  \\ 
        RAUNA \cite{RAUNA}  & 0.189 & 0.8848  & 19.00 \\
                RIRO \cite{RIRO}  & \it{0.144} &  0.9047 & 20.13 \\
                  URetinexNet++ \cite{uretinex++} & 0.151 & 0.8932  & 17.78 \\
                   Ours-Unfolding  & \bf{0.108} & \bf{0.9404}  & \bf{22.98} \\  
\hline
        Classical \\ \hline
  MSR \cite{msr-ipol}  & 0.442 & 0.7141 & 13.34 \\ 
  LIME \cite{LIME}  & 0.222 & 0.7855 & 14.28  \\ 
     LR3M \cite{LR3M} &  0.442 & 0.4654 & 4.11 \\  
  Structure-Retinex \cite{structure-Retinex} & 0.336  & 0.7103  & 13.21  \\       
Ours-Variational  &   \underline{0.143}  &  \it{0.9107} &  \it{21.28}  \\     
\end{tabular}
\caption{Quantitative evaluation on the LOLv1 test set \cite{deep-retinex}. Best results are highlighted in {\bf bold}, second best are \underline{underlined}, and third best in \textit{italic}. The proposed unfolding method achieves the best results for all metrics, while our variational model, despite not using learning strategies, is only outperformed by Retinexformer in PSNR and SSIM.}
\label{tabLOLv1}
\end{table}

To solve saddle-point optimization problems, the algorithm involves proximity operators, defined as $\prox_{\tau\phi}(x)=\arg\min_y \{\phi(y)+\tfrac{1}{2\tau}\|x-y\|_2^2\}$ for any proper convex function $\phi$. Since the energy term $\|R\circ L+N-\tilde{I}\|^2_{2}$ is not convex, we compute its proximity operator by minimizing the corresponding expression with respect to one variable while keeping the others fixed at their last updated values. This procedure is known as block coordinate descent and the steps lead to a local minimizer of the original energy \cite{chambolle2016introduction}. Since all terms involving $N$ are smooth, we can compute the exact minimum with respect to $N$ at each step. 

In the end, the primal-dual iterates are
\begin{equation}\label{primal-dual}
    \begin{array}{l}
    p^{n+1}_{k,i,j}= \dfrac{\alpha \big(p_{k,i,j}^n+\sigma (\nabla_\omega \overline{R}^{\,n})_{k,i,j}\big)}{\max\big(\alpha, \|p_{:,i,:}^n+\sigma (\nabla_{\omega}\overline{R}^{\,n})_{:,i,:}\|_2\big)},\\[2.5ex]
    q^{n+1}_{k,i,j,t}= \dfrac{\mu \left(q^n_{k,i,j,t}+\sigma\sqrt{\hat\omega_{i,j,t}}((\nabla \overline{R}^n)_{k,i,t}-(\nabla\hat{I})_{k,j,t})\right)}{\mu+\sigma},\\[2.5ex]
    o^{n+1}_{i,t}= \dfrac{\beta \big(o_{i,t}^n+\sigma (\nabla \overline{L}^{\,n})_{i,t}\big)}{\max\big(\beta, \|o_{i,:}^n+\sigma (\nabla\overline{L}^{\,n})_{i,:}\|_{2}\big)},\\[3.5ex]
   R^{n+1}=\dfrac{R^n+\tau \div_\omega p^{n+1}+\tau\widehat\div_{\hat\omega}q^{n+1}-\tau L^n(N^n-\tilde{I})}{1+\tau L^nL^n}, \\[2.5ex]
     \widetilde{R}_{k,i}^{n+1} = \max(0, \min(R_{k,i}^{n+1}, 1)),\\[2.5ex] \overline{R}^{n+1}=2\widetilde{R}^{n+1}-\widetilde{R}^n,  \\[2ex]
    L^{n+1}_i=\dfrac{L^n_i+\tau (\div o^{n+1})_i - \tau \sum_{k=1}^C R_{k,i}^{n+1} \left(N_{k,i}^n-\tilde{I}_{k,i}\right)}{1+\tau\sum_{k=1}^C R_{k,i}^{n+1}R_{k,i}^{n+1}}, \\[3ex]
    \widetilde{L}_i^{n+1} = \max(L_i^{n+1}, \max_k \tilde{I}_{k,i}),\\[2.5ex] \overline{L}^{n+1}=2\widetilde{L}^{n+1}-\widetilde{L}^n, \\[2ex]
    N^{n+1}=\dfrac{\tilde{I}-L^{n+1}R^{n+1}}{1+\lambda}.
    \end{array}
\end{equation}
In the above equations, $\div o\in\Rb^M$ is the classical divergence operator, $\div_{\omega}p\in\Rb^{C\times M}$ is defined in \eqref{eq-div-nltv}, and $\widehat{\div}_{\omega}q\in\Rb^{C\times M}$ is minus the adjoint operator of $(\nabla R-\nabla \hat{I})_{\hat{\omega}}$, defined as
\begin{equation*}
(\widehat\div_{\hat\omega}q)_{k,i}=\div\left(\sum_{j=1}^{(2\hat\nu+1)^2}\sqrt{\hat\omega_{i,j,t}} q_{k,i,j,t}\right).
\end{equation*}
Note that we have included in \eqref{primal-dual} the additional constraints $0\leq R\leq 1$ and $L\geq \tilde{I}$, which are standard assumptions in Retinex models \cite{kimmel}. The convergence of the resulting algorithm can be guaranteed in a manner similar to \cite{Ng}.

\subsection{Automatic gamma correction}\label{GammaIdeal}

Once the decomposition is complete, the next step is to adjust the illumination. A common approach is to apply gamma correction, which introduces an additional parameter $\gamma$ that must be empirically tuned for each image. Alternatively, based on the Gray-World assumption \cite{BUCHSBAUM19801}, we presume that the mean value of the enhanced illumination is approximately 0.5 and estimate $\gamma$ automatically.

Let $L\in \Rb^M$ be the illumination map obtained from our variational model. The mean value of the enhanced illumination with a $\gamma_0$ correction would be given by $\frac{1}{M}\sum_{i=1}^M L_i^{\gamma_0}.$ Since we want this value to be close to 0.5, our problem becomes finding a zero of the function $F(\gamma)={\frac{1}{M}}\sum_{i=1}^M L_i^{\gamma}-0.5$. To solve it, we  use the Newton-Raphson method, leading to the iterates
\begin{equation*}
    \gamma^{n+1}=\gamma^n-\frac{F(\gamma^n)}{F'(\gamma^n)}=\gamma^n-\frac{{ \frac{1}{M}}\sum_{i=1}^M L_i^{\gamma}-0.5}{ { \frac{1}{M}}\sum_{i=1}^M\left( L_i^{\gamma}\ln\left(L_i\right)\right)}.
\end{equation*}
In this way, we can efficiently estimate an appropriate value for $\gamma$ to adjust the illumination. Therefore, the final enhanced image is given by $I_{out}=L^\gamma R$.

\section{Deep Unfolding Twin}

In this section, we extend the proposed nonlocal Retinex-based variational model by introducing its deep unfolding counterpart, in which the proximal operators are replaced with learning-based networks, thereby avoiding the need of handcrafted priors. Specifically, the proximity operator for the reflectance component is substituted with a cross-attention residual network that mimics the behaviour of a nonlocal regularization. Inspired by SWIN transformers \cite{swir-transformer}, the proposed cross-attention mechanism modifies multi-head attention  \cite{multihead_ivan} to capture long-range dependencies by adjusting the keys and queries in each head. Moreover, the nonlocal gradient-type constraint is also reformulated using cross-attention.

\subsection{Algorithm unfolding}

We integrate two generic convex regularizers into the variational model \eqref{energia1}, leading to:
\begin{equation}\label{energia2}
\begin{aligned}
&\min_{R,L,N} \frac{1}{2}\|R\circ L+N-I\|_{2}^2+\alpha\Phi(R)+\beta \Psi(L)\\&+\frac{\lambda}{2} \|N\|_{2}^2 +\frac{\mu}{2} \|(\nabla R-\nabla I)_{\hat{\omega}}\|_{2}^2.
\end{aligned}
\end{equation}

Notice that all the terms, except the learnable regularizers, are differentiable. For this reason, we no longer need to dualize them, and thus, we choose a more suitable optimization technique for this case: the proximal gradient algorithm \cite{chambolle2016introduction}. The iterative scheme provided by this algorithm becomes
\begin{equation}\label{prox-grad}
    \begin{aligned}
    R^{n+1} &= \prox_{\tau \alpha \Phi} \left( \mathfrak{R}^{n} \right), \\
    L^{n+1}&=\prox_{\tau\beta\Psi}\left(\mathfrak{L}^{n}\right), \\
    N^{n+1} &= \dfrac{I - L^{n+1} R^{n+1}}{1 + \lambda},
    \end{aligned}
\end{equation}
where \begin{equation*}\begin{aligned}
\mathfrak{R}^{n}= R^n &- \tau \alpha L^n (R^n L^n + N^n - I)\\&+ \tau \alpha \mu \div \left( \nabla R^n - \sum_{j=1}^{(2 \hat{\nu}+1)^2} \hat{\omega}_j\nabla I_j \right),\end{aligned}\end{equation*}
\begin{equation*}\begin{aligned}
\mathfrak{L}^{n}= L^n-\tau \beta \sum_{k=1}^C R^{n+1}_k(L^n R_k^{n+1}+N_k-\left(I\right)_k).\end{aligned}\end{equation*}

In the rest of the paper, we refer to each step $n$ in the iterative scheme as a stage. The proximity operator related to the illumination regularization is replaced by a residual network ProxNet$^n$ while the proximity involved in the reflectance regularization is replaced by a cross-attention residual network CARNet$^n$.  
Moreover, we propose to replace the nonlocal operator of the gradient fidelity term by the cross-attention module  $\text{CA}^n_{\nabla R^n}$ to leverage the image self-similarity. 

Therefore, the unfolded version of the proximal gradient scheme results:
\begin{equation}
    \begin{aligned}
    R^{n+1} &= \text{CARNet}^{n+1}_I\left(\mathfrak{R}^{n} \right), \\
    L^{n+1}&=\text{ProxNet}^{n+1}\left(\mathfrak{L}^{n} \right), \\
    N^{n+1} &= \dfrac{I - L^{n+1} R^{n+1}}{1 + \lambda}.
    \end{aligned}
\end{equation}
That is, we divide each stage in three different steps, as illustrated in Figure \ref{fig:unfolding}. 
Now, $\mathfrak{R}^{n}$ is computed as
\begin{equation*}\begin{aligned}
\mathfrak{R}^{n}= R^n &- \tau \alpha L^n (R^n L^n + N^n - I)\\&+ \tau \alpha \mu \div \left( \nabla R^n - \text{CA}^n_{\nabla R^n}\left(\nabla I\right) \right).\end{aligned}\end{equation*}
The modules ProxNet$^n$, CARNet$^n_I$ and CA$^n_{\nabla R^n}$ do not share weights between the different stages but we maintain the same architectures in all stages. The proposed architectures are explained in detail in Subsection \ref{architecture}. Moreover, the hyperparameters $\alpha, \beta, \lambda, \mu$ and $\tau$ are shared across all stages and learned during the training phase.

In the first stage, $L^0$ and $R^0$ are initialized as $$L^0=\max_{k\in\{R,G,B\}}{I_k}, \hspace{0.2cm} R^0=\frac{I}{L+\varepsilon},$$  with $\varepsilon>0$ a small constant, while $N^0$ is initialized to all-zero. 

\subsection{Network architectures}\label{architecture}

The proximal operator can be expressed as follows
\begin{equation*}
\begin{aligned}
{y}=\prox_{\tau\phi}(x)&\iff x\in{y}+\tau\partial\phi\left({y}\right)\\&\iff{y}\in\left(Id+\tau\partial\phi\right)^{-1}\left(x\right).\end{aligned}
\end{equation*}
Therefore, the proximal operator can be interpreted as the inverse of a perturbation of the identity. From  this perspective, residual networks serve as good candidates for replacing the proximal operator in the unfolding framework. These networks consist of a convolutional neural network followed by
a skip connection, also known as a residual connection, making it an approach to identity. Then, we replace the function $\prox_{\tau\beta\Psi}$ by a residual network, ProxNet$^n$ presented in the $L$ step in Figure \ref{fig:unfolding}.

Building on the promising results of the variational model, we aim to continue exploiting self-similarities within the image to improve the estimation of the reflectance component. We propose the CARNet$^n_I$ network, which combines a residual network with a cross-attention module. Its architecture is illustrated in the R step of Figure \ref{fig:unfolding}. The cross-attention module is specifically designed to capture long-range dependencies of the image optimized for a high-performance on GPU. Each attention head focuses on different aspects of the input, approximating the nonlocal means filter:
$$NL(g)_i=\sum_j\omega_{i,j} g_j,$$
where $\omega_{i,j}$ are the classical nonlocal weights \eqref{classical_weights}.  

In the nonlocal networks introduced by Wang et al. \cite{nonlocal_learning}, the Euclidean distance between patches is replaced with the scalar product between pixels, while the filtering parameters are learned through convolutional operations. Nevertheless, this formulation requires a quadratic computational cost with respect to the number of pixels in the image. To mitigate this limitation, \cite{dosovitskiy2020image} proposes the use of a patch projection strategy, in which the image is represented as non-overlapping patches rather than individual pixels. As a result, each patch is replaced by a weighted average of the other patches.

Therefore, given an image $J\in \Rb^{C\times M}$,  we extract the non overlapping patches of size $S$ and by flattening them, we obtain $J_P=Proj_{S}(J)\in \Rb^{L\times T}$, with $L= C \cdot S\cdot S$ and $T = \frac{M}{S \cdot S}$. Then, the nonlocal filter, also called head-attention module, can be expressed as
\begin{equation*}
    \text{HA}(Q, K, V) = \text{Softmax}(W_q  Q \,\cdot\,W_k  K^T)\cdot W_v  V,
\end{equation*}
where $W_q, W_k$ and $W_v$ represent a linear operation and $Q$, $K$ and $V$ are the queries, keys and values. Specifically, \cite{dosovitskiy2020image} obtain the queries, keys and values applying a Layer Normalization (LN) and a Linear layer to the projected image $J_P$, i.e. $Q=K=V=\text{Linear}(\text{LN}(J_P))$. Finally, several self-attention layers are computed in parallel and fused with a Multi-Linear Perceptron in the so-called Multi-Head Attention. In this context, the role of the nonlocal weights is done by the operation between keys and queries, $\text{Softmax}(W_q  Q \,\cdot\,W_k  K^T)$, which can be computed using the input image itself, as in \cite{dosovitskiy2020image}, or alternatively, by replacing them with other auxiliary images.

The proposed cross-attention mechanism uses the input image as the values (the image to which the filter is applied) but computes the attention weights on different combination of keys and queries. In particular, we compute three head-attentions, the first computing the input image to obtain their long-range dependencies, the second with the observed image to capture the self-similarities of the pre-processed data, and the third that uses the input image as the keys but the observed image as the queries, obtaining the cross-attention relation between them.  Finally, to ensure a suitable performance and improve the efficiency in the attention computation, Instance Normalization (IN) and a Linear projection for reducing the patch representation are applied exclusively to the input keys and queries. Therefore, we have
\begin{align} \label{eq:CA}
\text{HA}_1 &= Up(HA(\widecheck{\mathfrak{R}}_P^n,\widecheck{\mathfrak{R}}_P^n, \mathfrak{R}_P^n),  \notag \\
\text{HA}_2 &= Up(HA(\widecheck{\mathfrak{R}}_P^n,\widecheck{J}_P, \mathfrak{R}_P^n),  \notag\\
\text{HA}_3 &= Up(HA(\widecheck{J}_P, \widecheck{J}_P, \mathfrak{R}_P^n),  \notag\\
\text{MHA} &= \text{MLP}([HA_1, HA_2, HA_3]),  \notag\\
\end{align}
where $Up$ is a pixel-shuffle operation that recovers the original image dimension, and $\widecheck{J}_P$ and $\widecheck{\mathfrak{R}}_P^n$ are obtained by applying IN and the Linear operator to the projection of $J$ and $\mathfrak{R}^n$. The corresponding parameters are shared neither among them nor between the different heads. Finally, following the approach in \cite{swir-transformer}, we repeat the proposed the cross-attention module in the proximity of the nonlocal regularizer by  previously shifting $S/2$ the resulting output and $J_P$ to avoid artifacts in the border of the image. We have indicated this process in Figure \ref{fig:unfolding} by Patch Cross-Attention (P-CA) and Shifted-Patch Cross-Attention (SP-CA).

Additionally, we adapt one cross-attention module to handle the nonlocal gradient-operators. To do this, we apply the same operation as in \eqref{eq:CA} but replacing the role of the $\mathfrak{R}_p^n$ by $(\nabla J)_p$ and $J_p$ by $\nabla R^n$. This approach provides to the unfolding framework the ability of our variational model to preserve fine details and edge structures.

\begin{figure*}[p]
    \centering
\begin{tabular}{c@{\hskip 0.2em} c@{\hskip 0.2em} c@{\hskip 0.2em} c@{\hskip 0.2em}}
\spyimage{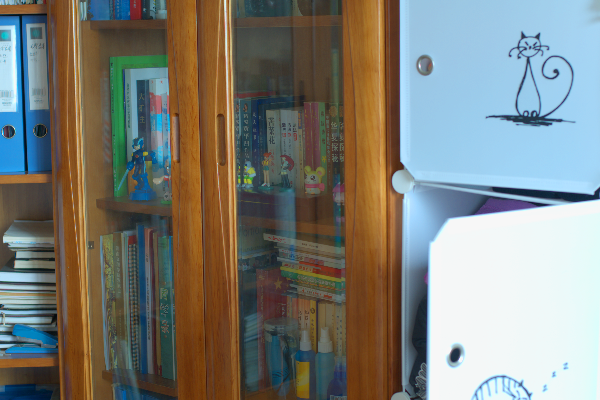} &\spyimage{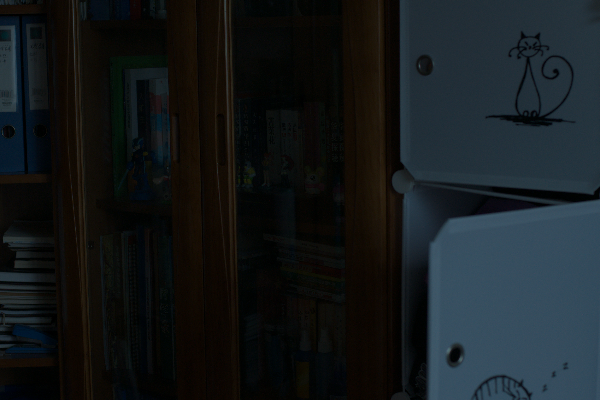}
    &
    \spyimage{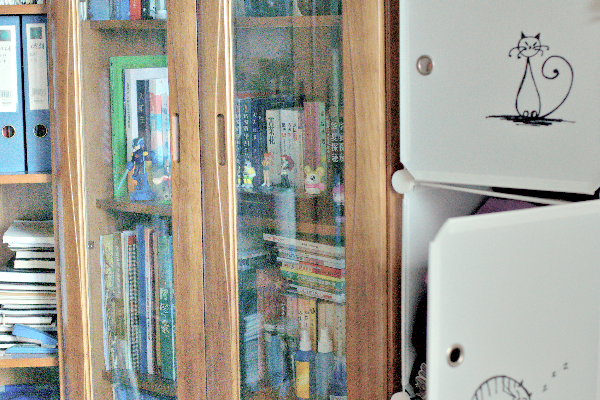} &
    \spyimage{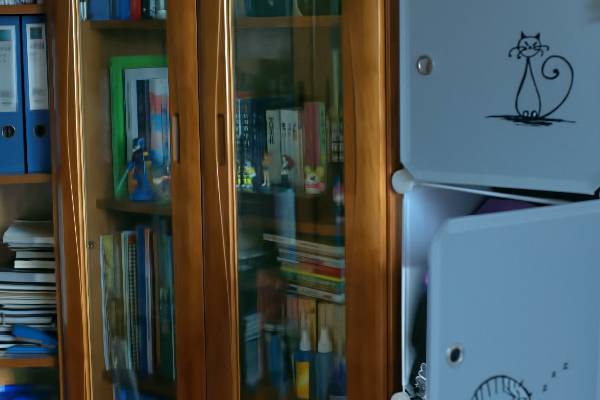} \\
     Ground Truth & Input &  MSR \cite{msr-ipol} & LIME \cite{LIME} \\ 
    \spyimage{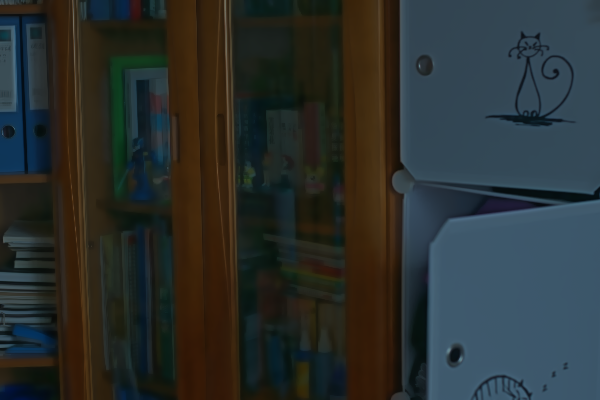} &\spyimage{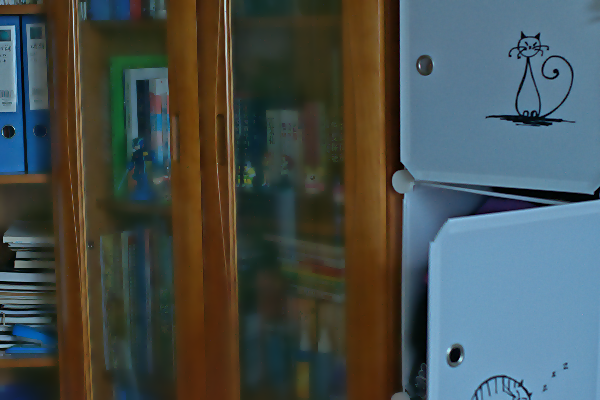}
    &
    \spyimage{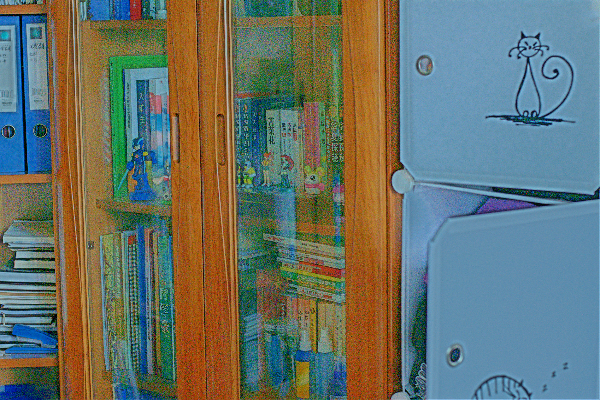} &
    \spyimage{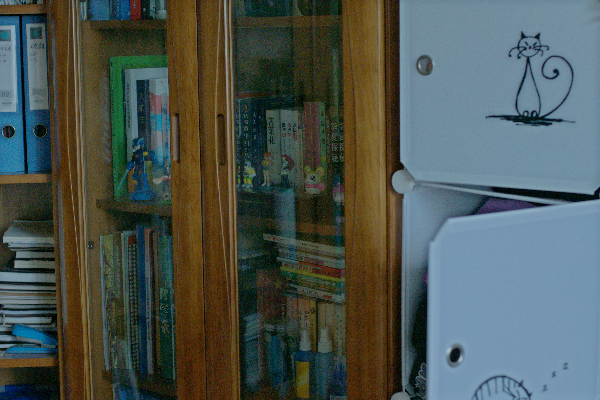} \\
    LR3M \cite{LR3M}  & Structure \cite{structure-Retinex} & RetinexNet \cite{deep-retinex} & ZeroDCE \cite{ZeroDCE} \\ \spyimage{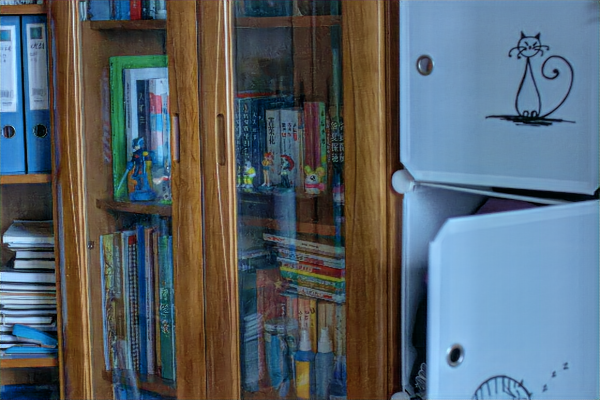} &\spyimage{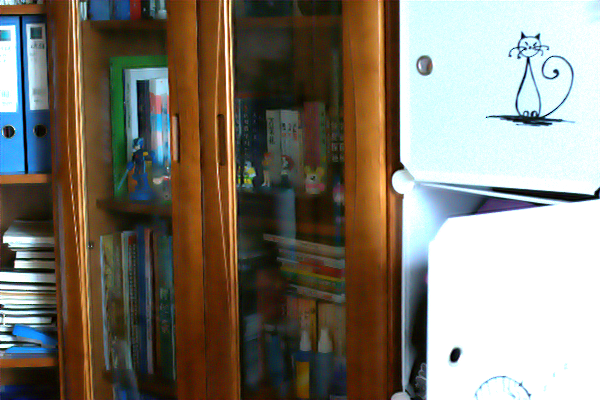}
    &
    \spyimage{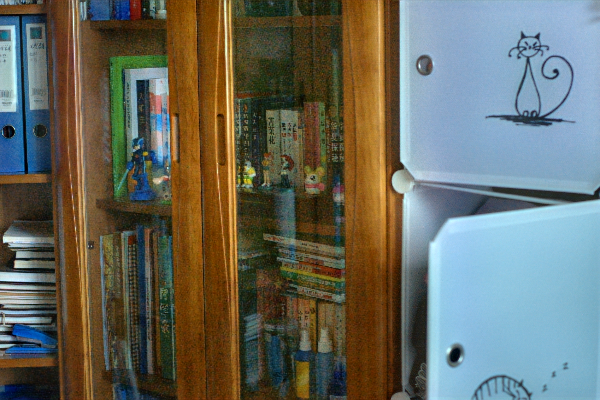} &
    \spyimage{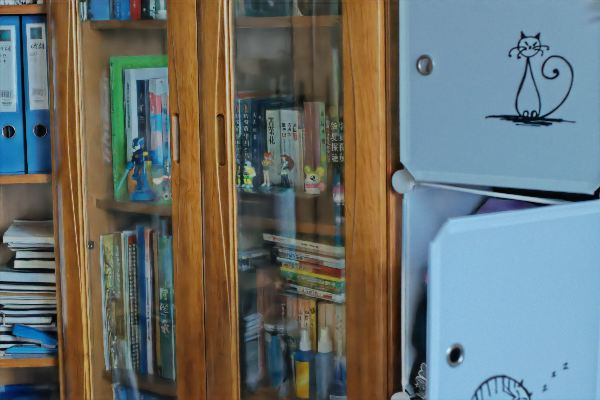} \\
    AGLLNet \cite{attention-guided} & RUAS \cite{RUAS} & EnlightenGAN \cite{enlightengan} & KinD+ \cite{KinDplus} \\ 
      \spyimage{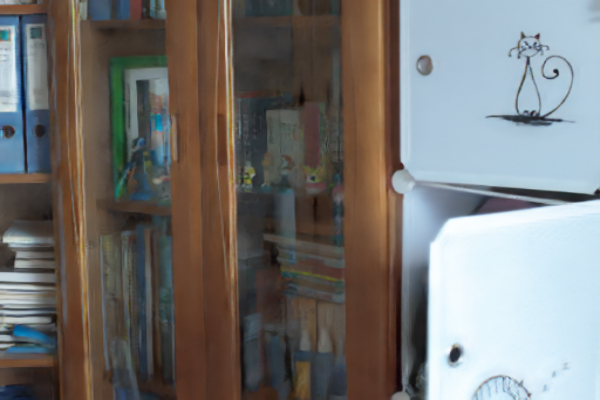} &
    \spyimage{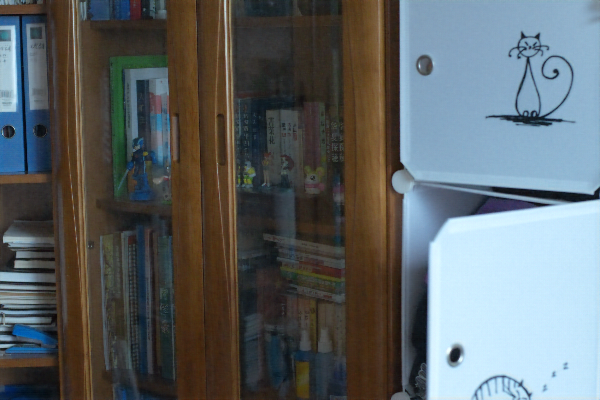} &
    \spyimage{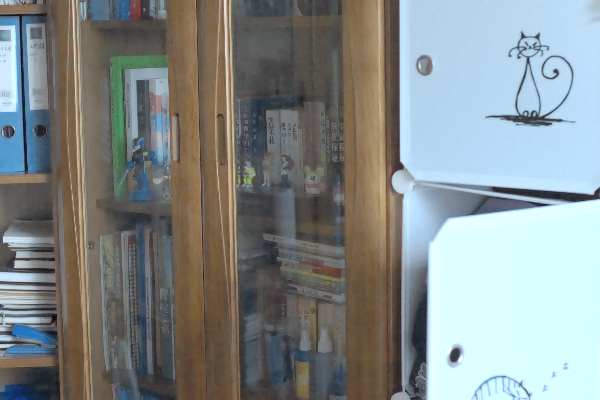} &
    \spyimage{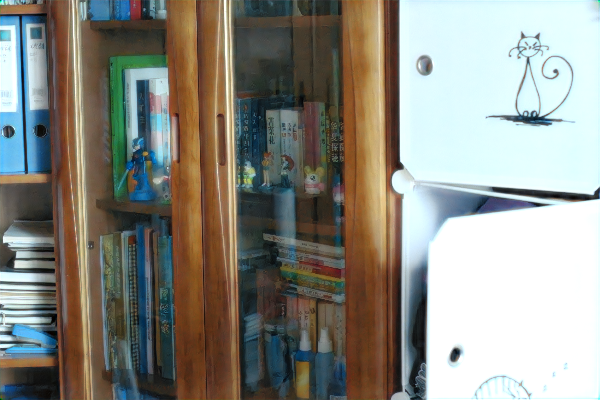} \\
    Night-Enhancement \cite{UnsupervisedNightImageEnhancement}  & Retinexformer \cite{RetinexFormer} & RAUNA \cite{RAUNA} & RIRO \cite {RIRO} \\
        \spyimage{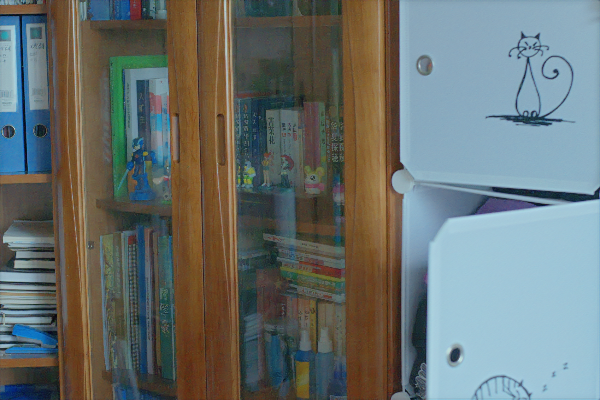} &
     \spyimage{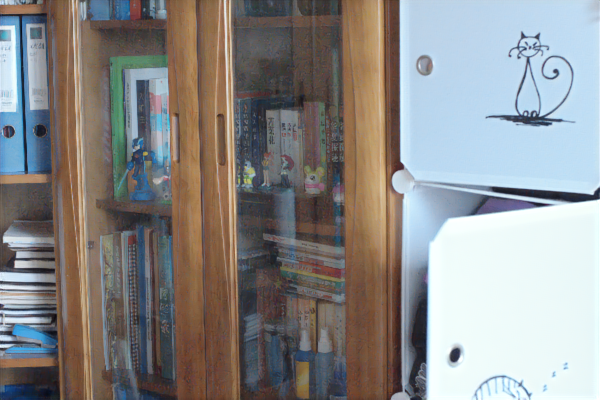} &
    \spyimage{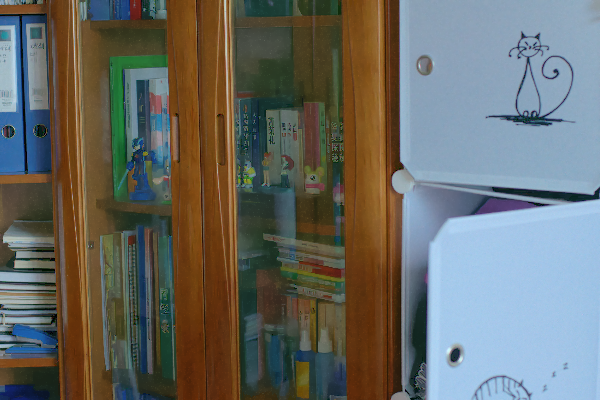} &
    \spyimage{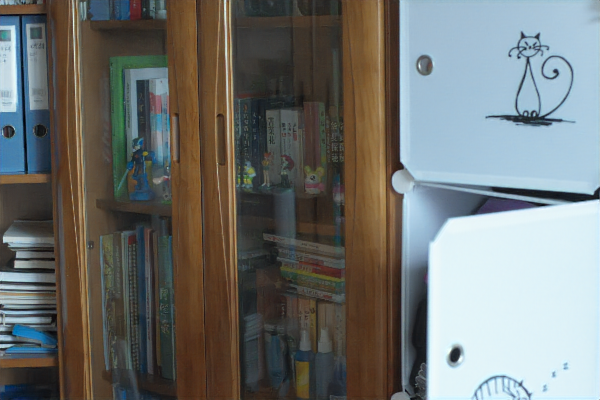} \\
     URetinexNet++ \cite{uretinex++} & Ghillie \cite{Ghillie} & Ours-Variational & Ours-Unfolding \\
\end{tabular}
    \caption{Visual comparison of the enhancement methods on a cropped image from the LOLv1 test set \cite{deep-retinex}. Only URetinexNet++, KinD+, and the two proposed models successfully enhance the image and reduce noise without causing oversaturation or excessive smoothing. However, as observed in the zoomed-in region, some noise remains in URetinexNet++, while KinD+ tends to overemphasize details, resulting in unnatural textures.}
    \label{fig:LOLcomparison}
\end{figure*}

\begin{figure*}[p]
    \centering
\begin{tabular}{c@{\hskip 0.2em} c@{\hskip 0.2em} c@{\hskip 0.2em} c}
\spyimagelol{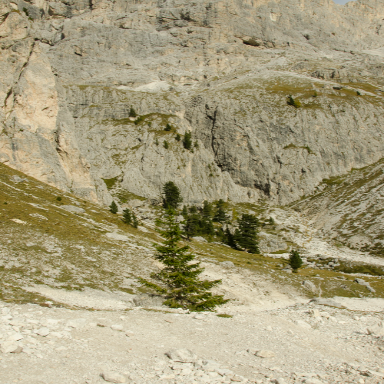} &\spyimagelol{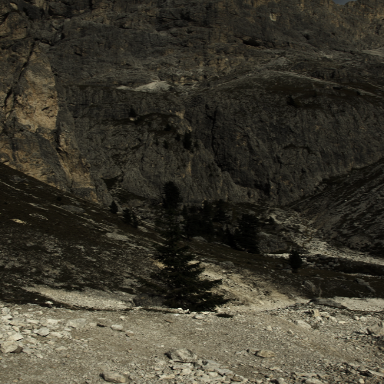}
    &
    \spyimagelol{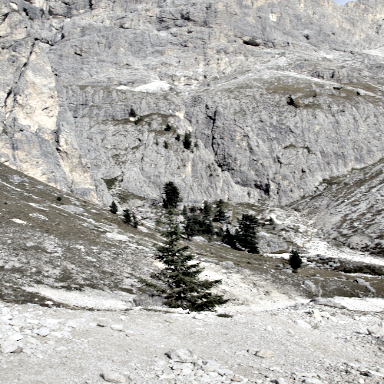} &
    \spyimagelol{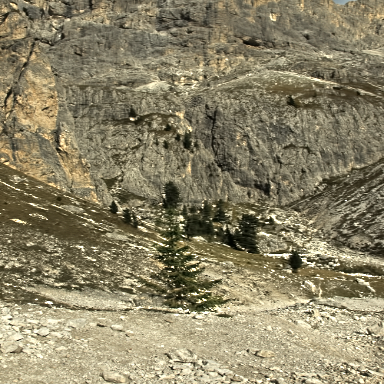} \\
     Ground Truth & Input &  MSR \cite{msr-ipol} & LIME \cite{LIME} \\ 
    \spyimagelol{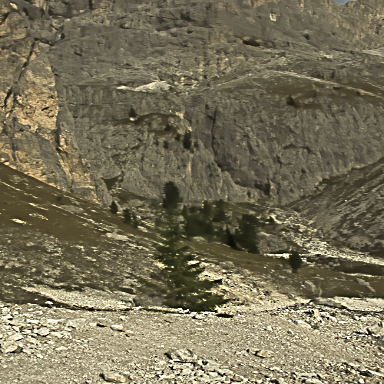} &\spyimagelol{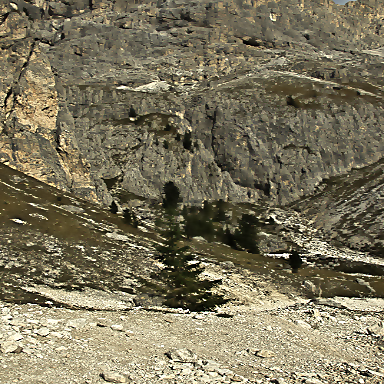}
    &
    \spyimagelol{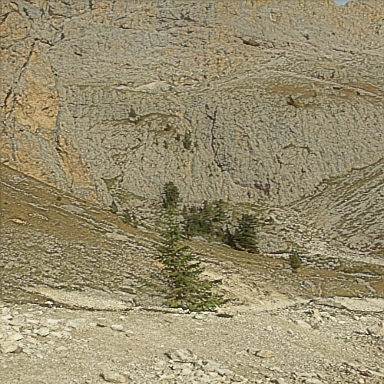} &
    \spyimagelol{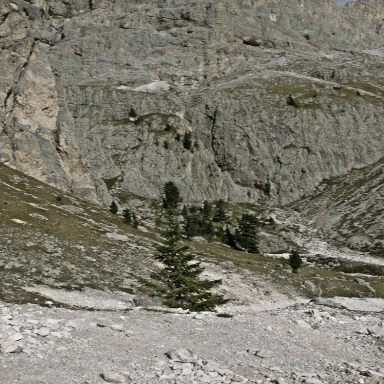} \\
    LR3M \cite{LR3M} & Structure \cite{structure-Retinex} & RetinexNet \cite{deep-retinex} & ZeroDCE \cite{ZeroDCE} \\ \spyimagelol{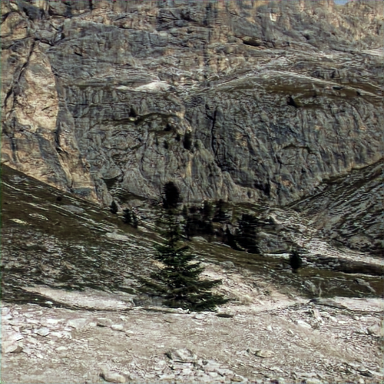} &\spyimagelol{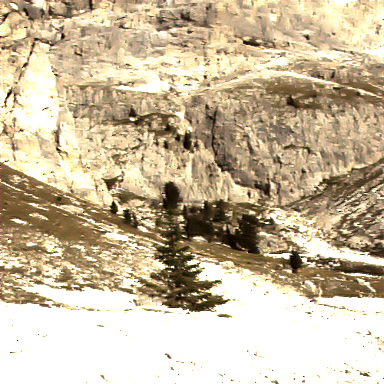}
    &
    \spyimagelol{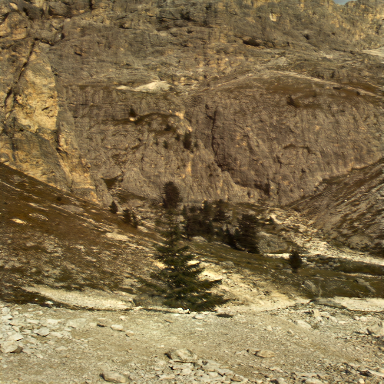} &
    \spyimagelol{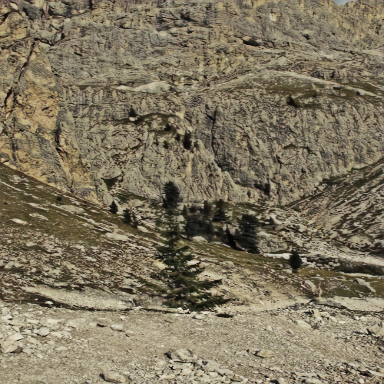} \\
    AGLLNet \cite{attention-guided}  & RUAS \cite{RUAS} & EnlightenGAN \cite{enlightengan}& KinD+ \cite{KinDplus} \\ 
      \spyimagelol{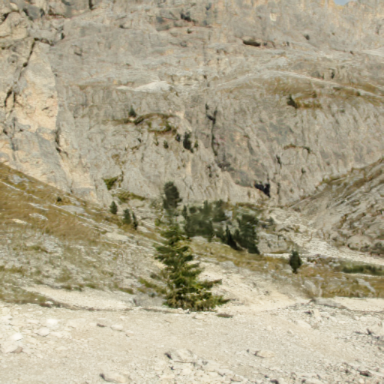} &
    \spyimagelol{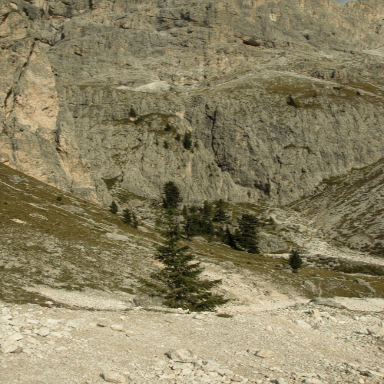} &
    \spyimagelol{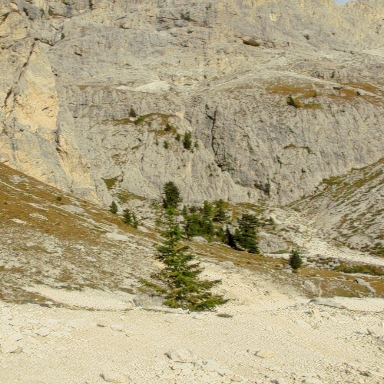} &
    \spyimagelol{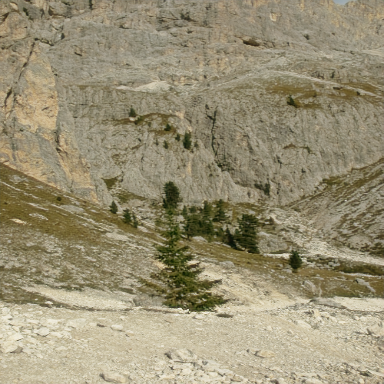} \\
    Night-Enhancement \cite{UnsupervisedNightImageEnhancement} & Retinexformer \cite{RetinexFormer} & RAUNA \cite{RAUNA} & RIRO \cite{RIRO} \\
        \spyimagelol{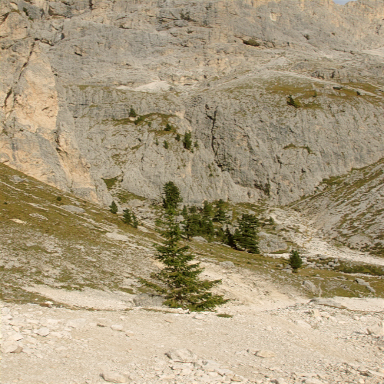} &
     \spyimagelol{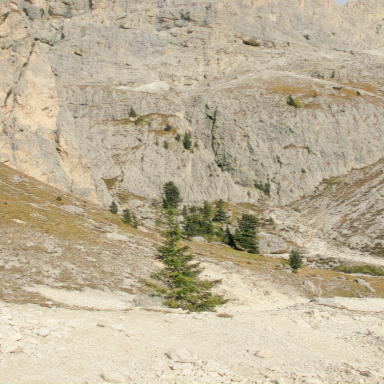} &
    \spyimagelol{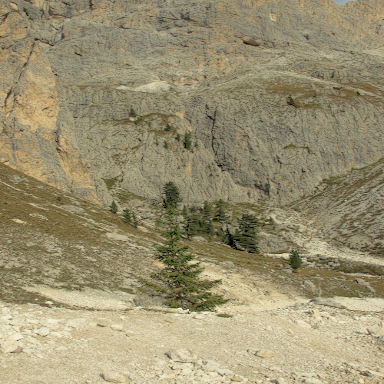} &
    \spyimagelol{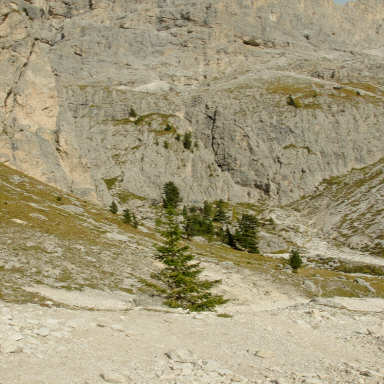} \\
     URetinexNet++ \cite{uretinex++} & Ghillie \cite{Ghillie}  & Ours-Variational & Ours-Unfolding \\
\end{tabular}
    \caption{Visual comparison of the enhancement methods on a cropped image from the LOLv2-Synthetic test set \cite{sparse}. Retinexformer, RAUNA, RIRO, URetinexNet++, and our two methods produce satisfactory results. However, considering the color restoration across different objects in the scene, our unfolding approach demonstrates the most effective performance.}
    \label{fig:LOLv2comparison}
\end{figure*}

\section{Experimental Results}

In this section, we assess the performance of the proposed method for low-light image enhancement and compare it with state-of-the-art techniques. 
We use the LOLv1 \cite{deep-retinex} and LOLv2 \cite{sparse} as reference datasets. Additionally, we display results on real low-light images from the non-reference LIME dataset \cite{LIME}.

We compare the proposed approach with the classical methods MSR \cite{msr-ipol}, LIME \cite{LIME}, LR3M \cite{LR3M}, and Structure-Retinex \cite{structure-Retinex}; the purely deep learning techniques RetinexNet \cite{deep-retinex}, ZeroDCE \cite{ZeroDCE}, AGLLNet \cite{attention-guided}, EnlightenGAN\cite{enlightengan}, KinD+ \cite{KinDplus}, Night-Enhancement \cite{UnsupervisedNightImageEnhancement}, Retinexformer \cite{RetinexFormer}, and Ghillie \cite{Ghillie}; and the unfolding methods RUAS \cite{RUAS}, RAUNA \cite{RAUNA}, RIRO \cite{RIRO} and URetinexNet++ \cite{uretinex++}. LIME was implemented by us, while the source codes of all other methods were obtained from the authors' webpages. All trainings were performed according to the specified configurations.

In the context of computer vision, measuring the similarity between images is a challenging task, especially in the low-light image enhancement context, where several factors such as illumination adjustment, noise suppression, contrast augmentation, and color correction must be taken into account. LPIPS \cite{LPIPS} has been shown to be a perceptual metric closely aligned with human visual perception.
Therefore, we empirically optimized the parameters of our variational model based on it. Additionally, we include PSNR (Peak Signal to Noise Ratio) \cite{rabbani1991digital}, which assesses the spatial reconstruction quality with respect to noise and SSIM (Structural Similarity Index Measure) \cite{wang2002universal}, which evaluates the overall quality of the enhanced image, as quality metrics.

Due to the absence of ground-truth in the LIME dataset, we assess the performance using NIQE (Natural Image Quality Evaluator) \cite{NIQE}, a non-reference metric that evaluates quality based on a model derived from the statistical features of natural scenes. 

The proposed deep unfolding network is trained  in an end-to-end manner over 1000 epochs using the loss function
\begin{equation*}
    \begin{aligned}
\text{Loss}\left(I_{out}, I_{gt}\right)&\hspace{-0.08cm}=\hspace{-0.08cm}\text{MSE}\left(I_{out}, I_{gt}\right)\hspace{-0.05cm}+\hspace{-0.05cm}\alpha_1 \text{Loss}_c\hspace{-0.05cm}\left(I_{out}, I_{gt}\right)\\&\hspace{-0.05cm}+\hspace{-0.05cm}\alpha_2 \text{LPIPS}\left(I_{out}, I_{gt}\right),    \end{aligned}
\end{equation*}
where $I_{gt}$ represents the ground-truth image, $\alpha_1$ and $\alpha_2$ are fixed to 0.1, MSE denotes the mean squared error, and $\text{Loss}_c$ \cite{color_loss} minimizes the cosine distance between the true and predicted values to reduce the color degradation as follows:
$$\text{Loss}_c\hspace{-0.05cm}\left(I_{out}, I_{gt}\right)=
\frac{\sum_{i=1}^M\sum_{k=1}^C \left(I_{out}\right)_{k,i}\left(I_{gt}\right)_{k,i}}{MC}.$$
We use Adam optimizer with an initial learning rate of $10^{-4}$ and set the number of primal-dual stages to 5, since we have experimentally checked that this is an optimal value.

\subsection{Experiments on LOLv1 dataset}

The LOLv1 dataset \cite{deep-retinex} consists of 500 pairs of low/normal-light images, capturing a diverse range of real-world scenes under different exposure conditions. For evaluation purposes, the dataset is divided into a training set of 485 image pairs and a test set of 15 image pairs. 

Table \ref{tabLOLv1} displays the quantitative metrics obtained for each technique on the test set. The proposed unfolding methods outperforms all competing approaches under all evaluation metrics. Additionally, our variational model ranks second in LPIPS and third in PSNR and SSIM, surpassed only by Retinexformer.

Figure \ref{fig:LOLcomparison} shows crops of the enhanced results produced by each method on a sample from the test set. MSR, RetinexNet, Zero-DCE, EnlightenGAN, and URetinexNet++ are unable to effectively remove noise, while methods such as LIME, RUAS, and Night-Enhancement produce over-smoothed results. Moreover, illumination adjustment issues are evident, either due to excessive brightness, as in RIRO, RUAS and MSR, or insufficient enhancement, as in LR3M, Structure-Retinex, and Zero-DCE. Additionally, some methods, like AGLLNet and KinD+, generate unnatural textures, while others, such as RAUNA and Ghillie, result in considerable color loss. Retinexformer and our unfolding technique also experience slight color degradation, but this issue is not present in our variational approach.

\subsection{Experiments on LOLv2-Synthetic dataset}

The LOLv2 dataset \cite{sparse} is divided into two subsets: Real and Synthetic. LOLv2-Real contains images captured under similar conditions to those in LOLv1, while LOLv2-Synthetic is composed of high-quality RAW images that have been processed to simulate low-light conditions. Therefore, we have conducted the evaluation on the Synthetic subset to assess the performance of the methods on a different type of data. The LOLv2-Synthetic dataset is divided into 900 image pairs for training and 100 image pairs for testing.

All deep learning and unfolding techniques have been retrained on this dataset. However, for computational purposes, the parameters involved in our variational model have only been slightly modified from their optimal values obtained on the LOLv1 dataset by optimizing on a small subset of possible combinations. This has evidently limited its performance.

Table \ref{tabLOLv2} displays the average metrics on the test set. The proposed unfolding method ranks first in LPIPS and second in SSIM and PSNR, with only RetinexFormer outperforming it. Our variational model, despite its parameters not being accurately optimized, achieves competitive results, ranking third in terms of PSNR.

\begin{table}[t]
\centering
\begin{tabular}{lccc}
  Method   & LPIPS $\downarrow$ & SSIM $\uparrow$ & PSNR $\uparrow$\\ \hline
        Pure deep learning \\ \hline
   RetinexNet \cite{deep-retinex}  & 0.262  & 0.8413  & 18.87 \\
       ZeroDCE \cite{ZeroDCE} & 0.168 & 0.8406  & 17.74 \\
AGLLNet \cite{attention-guided} & 0.234 & 0.8125  & 16.82 \\
EnlightenGAN \cite{enlightengan}  & 0.212 &  0.8055 & 16.53 \\ 
        KinD+ \cite{KinDplus}  & 0.231 & 0.7799 & 16.90 \\ 
                Night-Enhancement \cite{UnsupervisedNightImageEnhancement} & 0.193 & 0. 8356 & 22.55 \\
        Retinexformer \cite{RetinexFormer}  & \underline{0.064} &  \bf{0.9503} & \bf{24.76} \\
        Ghillie \cite{Ghillie}  & 0.144 & 0.8759  & 18.38 \\
        \hline 
        Unfolding\\ \hline
          RUAS \cite{RUAS} & 0.361 & 0.6644  & 13.81  \\ 
        RAUNA \cite{RAUNA}  & 0.118 & 0.8790 & 20.44  \\
                RIRO \cite{RIRO}  & 0.109  &  0.9110 &  20.85 \\
                  URetinexNet++ \cite{uretinex++} & \it{0.085}  &  \it{0.9282} & 21.81  \\
                    Ours-Unfolding & \bf{0.062} & \underline{0.9457}  & \underline{24.39}  \\
\hline       Classical \\ \hline
  MSR \cite{msr-ipol}  & 0.238 & 0.8153 & 16.37 \\ 
  LIME \cite{LIME}  & 0.214 & 0.8212 &  17.67 \\ 
     LR3M \cite{LR3M} & 0.327  & 0.7156 & 16.71 \\  
  Structure-Retinex \cite{structure-Retinex} & 0.338 & 0.6754  &  16.16 \\       
Ours-Variational  &  0.114   & 0.9119  & \it{22.78}  \\     
\end{tabular}
\caption{Quantitative evaluation on the LOLv2-Synthetic test set \cite{sparse}. Best results are highlighted in {\bf bold}, second best are \underline{underlined}, and third best in \textit{italic}. The proposed unfolding method ranks first in LPIPS and second in SSIM and PSNR. Our variational model, despite its parameters being only slightly modified from the values from the LOLv1 dataset and not being accurately optimized as done for all deep learning and unfolding approaches, achieves competitive results.}
\label{tabLOLv2}
\end{table}

\begin{figure*}[p]
    \centering
\begin{tabular}{c@{\hskip 0.2em} c@{\hskip 0.2em} c@{\hskip 0.2em} c} &\spyimagelamp{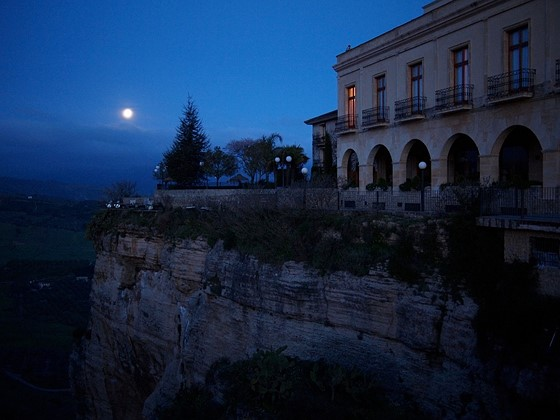}
    &
    \spyimagelamp{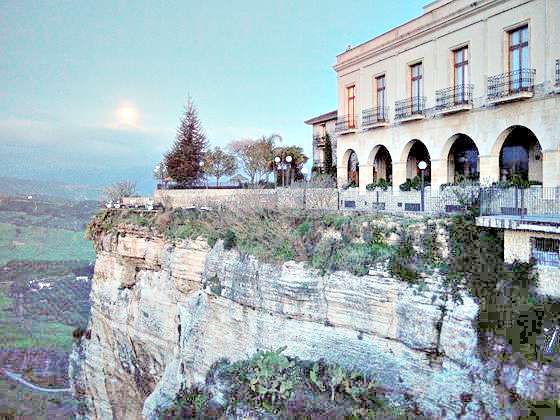} &
    \spyimagelamp{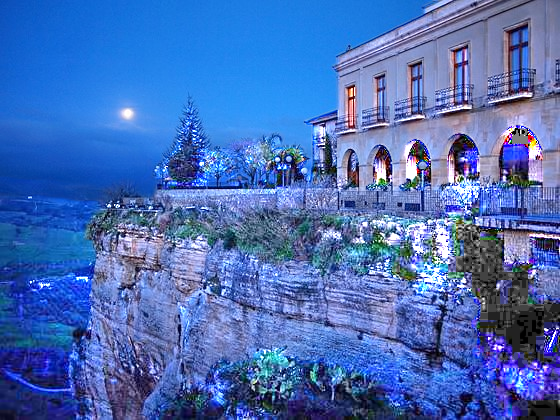} \\
      & Input &  MSR \cite{msr-ipol} & LIME \cite{LIME} \\ 
    \spyimagelamp{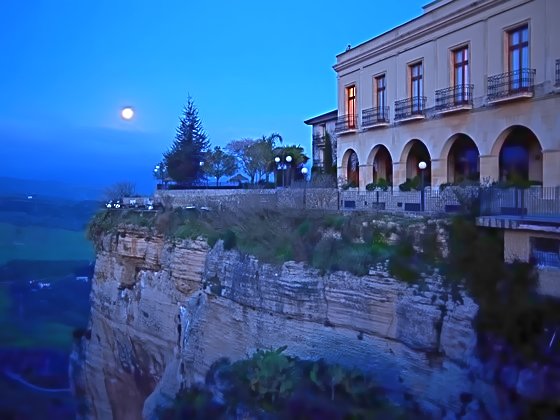} &\spyimagelamp{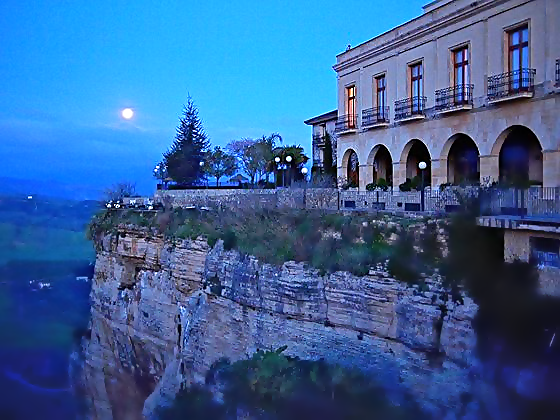}
    &
    \spyimagelamp{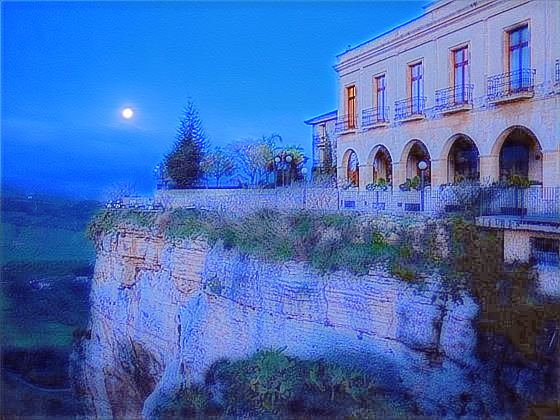} &
    \spyimagelamp{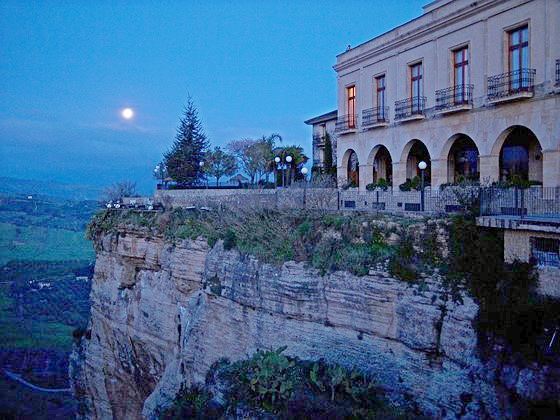} \\
    LR3M \cite{LR3M}  & Structure \cite{structure-Retinex} & RetinexNet \cite{deep-retinex} & ZeroDCE \cite{ZeroDCE} \\ \spyimagelamp{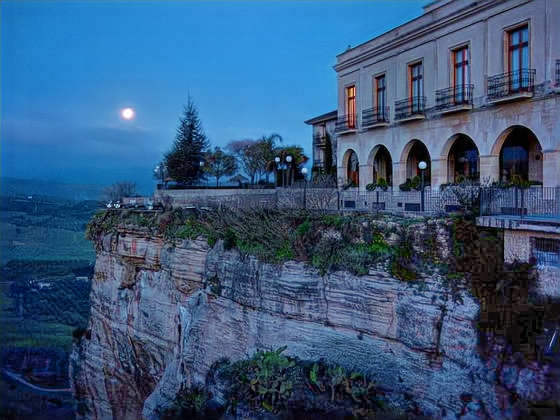} &\spyimagelamp{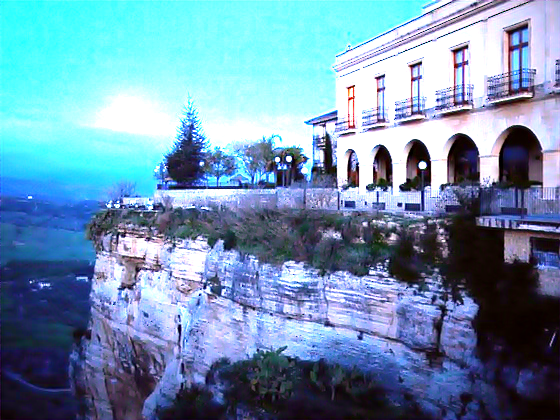}
    &
    \spyimagelamp{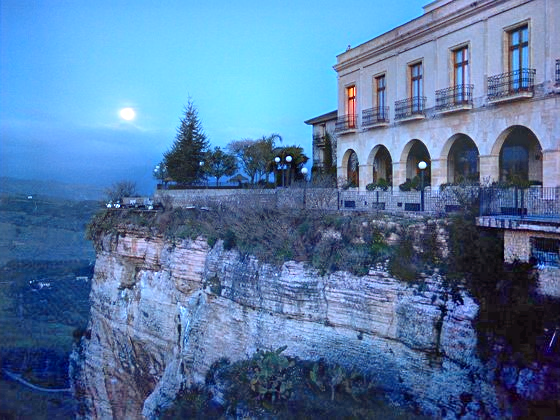} &
    \spyimagelamp{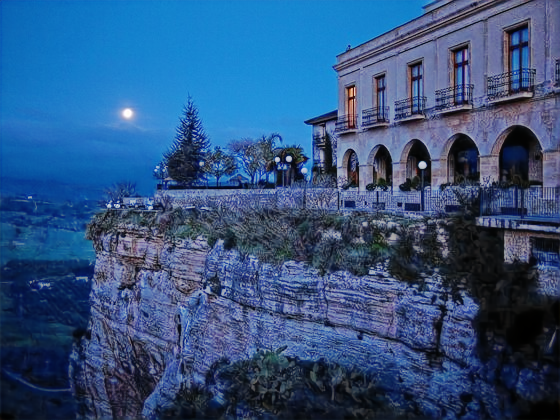} \\
    AGLLNet \cite{attention-guided} & RUAS \cite{RUAS} & EnlightenGAN \cite{enlightengan} & KinD+ \cite{KinDplus} \\ 
      \spyimagelamp{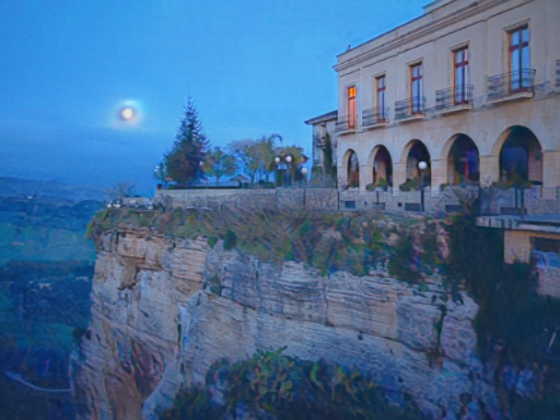} &
    \spyimagelamp{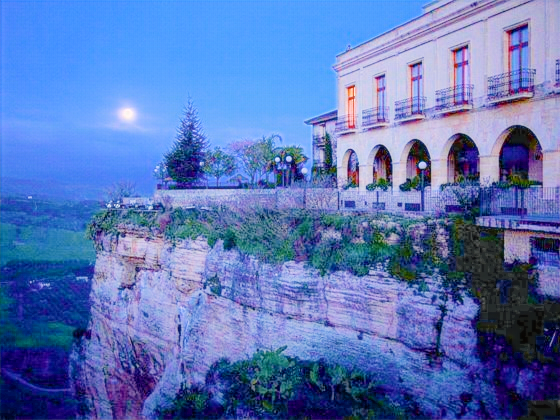} &
    \spyimagelamp{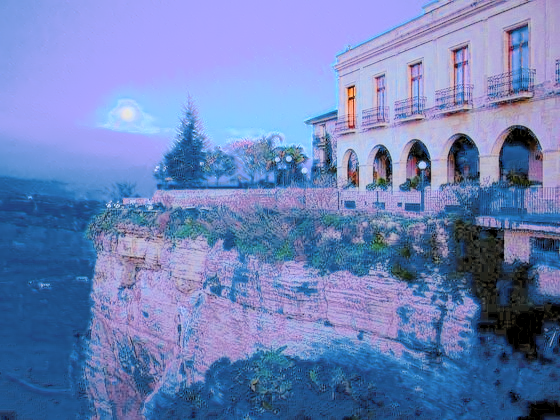} &
    \spyimagelamp{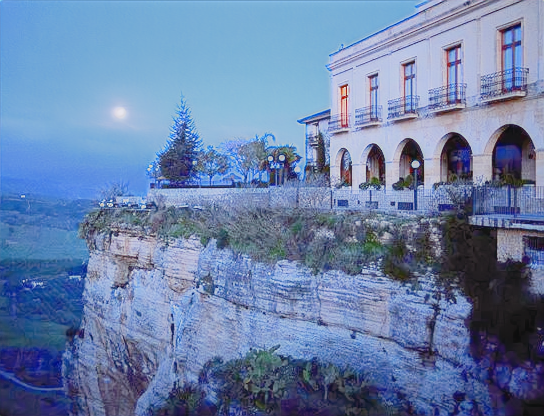} \\
    Night-Enhancement \cite{UnsupervisedNightImageEnhancement} & Retinexformer \cite{RetinexFormer} & RAUNA \cite{RAUNA} & RIRO \cite{RIRO} \\
        \spyimagelamp{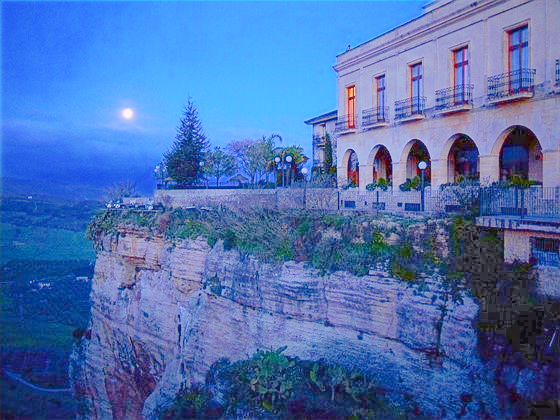} &
     \spyimagelamp{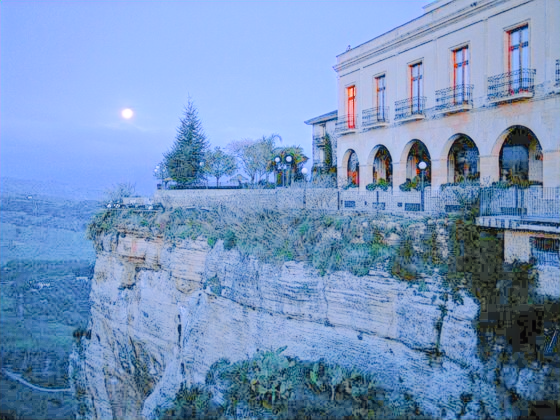} &
    \spyimagelamp{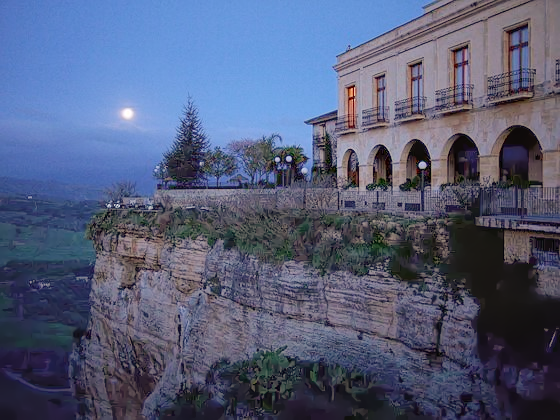} &
    \spyimagelamp{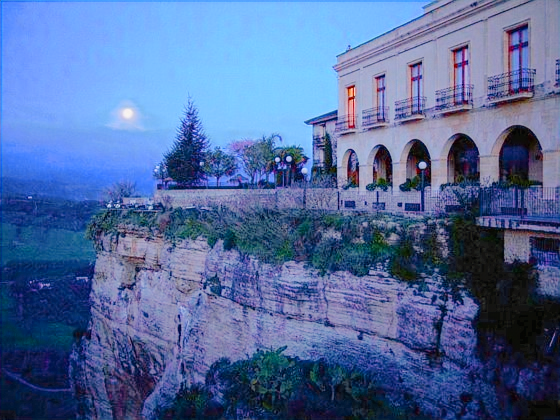} \\
     URetinexNet++ \cite{uretinex++} & Ghillie \cite{Ghillie}  & Ours-Variational & Ours-Unfolding \\
\end{tabular}
    \caption{Visual comparison on LIME image. Several methods like LIME and RetinexNet introduce visible artifacts, while others like LR3M and RAUNA create halos around the edges. In contrast, our results show high-quality results, with the variational approach producing more realistic colors.}
    \label{fig:LIMEcomparison}
\end{figure*}

Figure \ref{fig:LOLv2comparison} shows crops of the enhanced images on a sample of the LOLv2-Synthetic test set. The results generally exhibit minimal noise. However, some methods fail to recover textures, generating artificial patterns in the case of RetinexNet, while LR3M, RUAS, and Night-Enhancement produce excessively smoothed images that lose important details. Nevertheless, the most significant difference is in color restoration. Retinexformer, RAUNA, RIRO, URetinexNet++, and our approaches effectively mitigate color distortions, producing satisfactory results, but our unfolding model achieves the most faithful color reconstruction, especially in elements such as the tree and rocks.

\begin{figure*}[t]
\begin{minipage}[b]{.24\linewidth}
  \centering
  \centerline{\includegraphics[width=\textwidth]{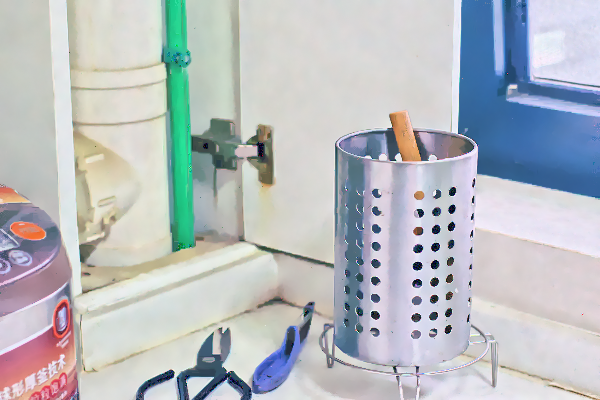}}
  \centerline{Reflectance}
\end{minipage}
\hfill
\begin{minipage}[b]{0.24\linewidth}
  \centering
  \centerline{\includegraphics[width=\textwidth]{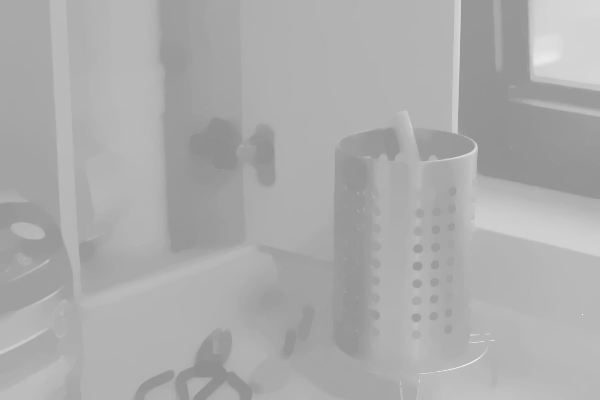}}
  \centerline{Enhanced illumination}
\end{minipage}
\hfill
\begin{minipage}[b]{.24\linewidth}
  \centering
  \centerline{\includegraphics[width=\textwidth]{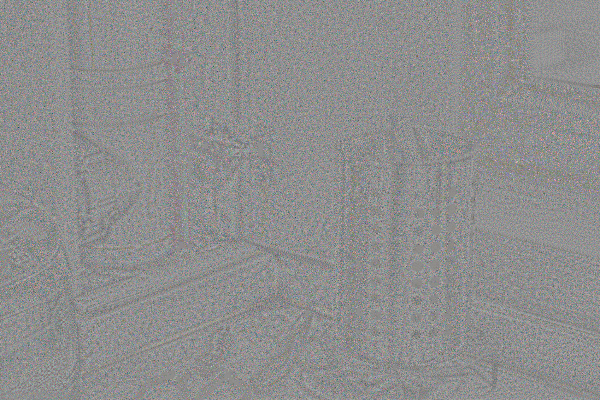}}
  \centerline{Noise}
\end{minipage}
\hfill
\begin{minipage}[b]{0.24\linewidth}
  \centering
  \centerline{\includegraphics[width=\textwidth]{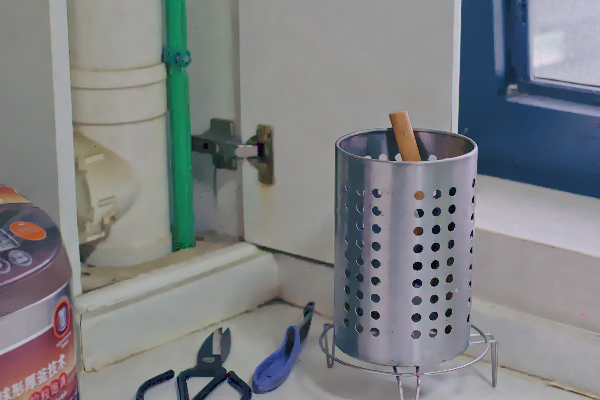}}
  \centerline{Output}
\end{minipage}
\hfill
\vspace{0.1cm}
\caption{Decomposition results of the proposed low-light image enhancement variational model. The reflectance contains geometry and texture, the illumination captures light intensity, and the noise component does not retain significant structural details.}
\label{fig:decomposition}
\end{figure*}

\subsection{Experiments on LIME dataset}

The LIME dataset \cite{LIME} includes 10 natural images captured in low-light conditions, without ground-truth references. It is especially significant for evaluating the generalization capabilities of enhancement methods in real-world scenarios, where paired low-light and high-light images are typically not available.

\begin{table}[t]
\centering
\begin{tabular}{lc}
  Method   &  NIQE $\downarrow$\\ \hline
        Purely deep learning \\ \hline
   RetinexNet \cite{deep-retinex} & 5.9355 \\
       ZeroDCE \cite{ZeroDCE} & 3.9695 \\
AGLLNet \cite{attention-guided} &  4.4824\\
EnlightenGAN \cite{enlightengan}  & \underline{3.5276} \\ 
        KinD+ \cite{KinDplus}  & 4.7610\\ 
        Night-Enhancement \cite{UnsupervisedNightImageEnhancement} & 4.0420\\
        Retinexformer \cite{RetinexFormer} & 3.7646 \\
        Ghillie \cite{Ghillie}& 3.7807 \\
        \hline 
        Unfolding\\ \hline
          RUAS \cite{RUAS} & 4.5186 \\ 
        RAUNA \cite{RAUNA}  & 4.4347  \\
                RIRO \cite{RIRO}  & 4.4393 \\
                  URetinexNet++ \cite{uretinex++} & 3.9023 \\
                    Ours-Unfolding &\it{3.5649} \\ 
\hline
        Classical \\ \hline
  MSR \cite{msr-ipol}  & 3.9016  \\    
  LIME \cite{LIME}  & 4.5211  \\
     LR3M \cite{LR3M} & 4.4000 \\        
  Structure-Retinex \cite{structure-Retinex} & 4.5539 \\ 
    Ours-Variational  & \bf{3.4136} \\ 
\end{tabular}
\caption{Quantitative evaluation on the non-reference LIME dataset \cite{LIME}. Best results are highlighted in {\bf bold}, second best are \underline{underlined}, and third best in \textit{italic}. The proposed variational model achieves the best results, while its unfolding counterpart ranks third, being outperformed by EnlightenGAN.}
\label{tabLIME}
\end{table}

Table \ref{tabLIME} shows that our variational method achieves the best NIQE value, while the unfolding approach is only outperformed by EnlightenGAN. As illustrated in Figure \ref{fig:LIMEcomparison}, several methods, including LIME, RetinexNet, Night-Enhancement, and Retinexformer, produce visible artifacts that significantly degrade the overall quality of the enhanced images. Other methods like LR3M, Structure, RIRO, and RAUNA introduce undesirable halos around the edges of the objects. Furthermore, MSR and RUAS fail to achieve accurate lighting conditions in the scene. In contrast, both of our results exhibit significant improvements in these areas. Again, the variational approach produces more realistic colors compared to all other methods, including its unfolding counterpart, making it more faithful to the real-world lighting context.

\begin{figure}[t]
\begin{minipage}[b]{.48\linewidth}
  \centering
  \centerline{\adjustbox{trim={.2\width} {0.4\height} {0.45\width} {0.2\height},clip, width=\textwidth}%
  {\includegraphics[width=\textwidth]{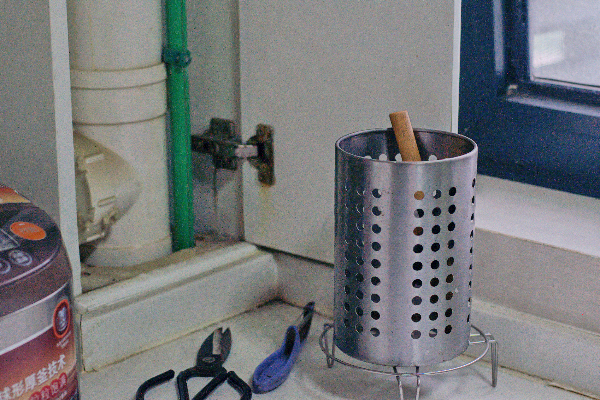}}}
  \centerline{Output without $N$}
\end{minipage}
\hfill
\begin{minipage}[b]{0.48\linewidth}
  \centering
  \centerline{\adjustbox{trim={.2\width} {0.4\height} {0.45\width} {0.2\height},clip, width=\textwidth}%
  {\includegraphics[width=\textwidth]{111Ours.png}}}
  \centerline{Output with $N$}
\end{minipage}
\caption{Influence of the noise component $N$ in \eqref{energia1}, which is crucial to prevent the enhanced image from being noisy.}
\label{fig:noise_term}
\end{figure}

\begin{figure}[t]
\begin{minipage}[b]{.48\linewidth}
  \centering
  \centerline{\adjustbox{trim={.1\width} {0.32\height} {0.68\width} {0.45\height},clip, width=\textwidth}%
  {\includegraphics[width=\textwidth]{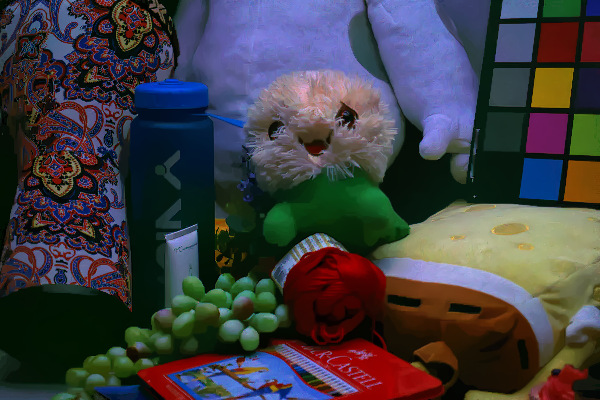}}}
 \parbox{\textwidth}{\centering No color correction}
\end{minipage}
\hfill
\begin{minipage}[b]{.48\linewidth}
  \centering
      \vfill
 \centerline{\adjustbox{trim={.1\width} {0.32\height} {0.68\width} {0.45\height},clip, width=\textwidth}%
  {\includegraphics[width=\textwidth]{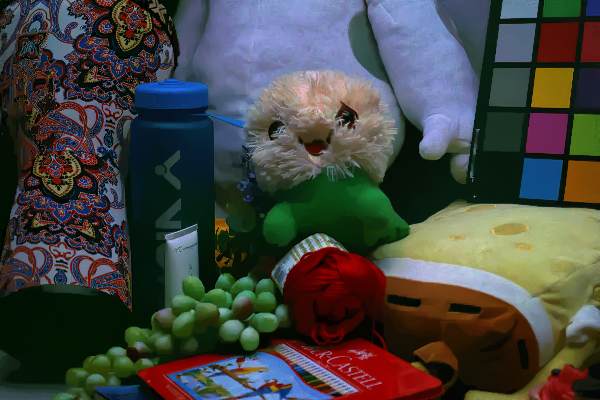}}}
  \parbox{\textwidth}{\centering Color correction}
\end{minipage}
\caption{Impact of color correction on the low-light image. The final result shows objects with a purer white tone, as the predominance of the blue channel has been eliminated.}
\label{fig:color}
\end{figure}

\begin{figure*}[t]
\hfill\begin{minipage}[b]{.32\linewidth}
  \centering
  \centerline{\adjustbox{trim={0.15\width} {0.1\height} {0.1\width} {0.1\height},clip, width=\textwidth}%
  {\includegraphics[width=\textwidth]{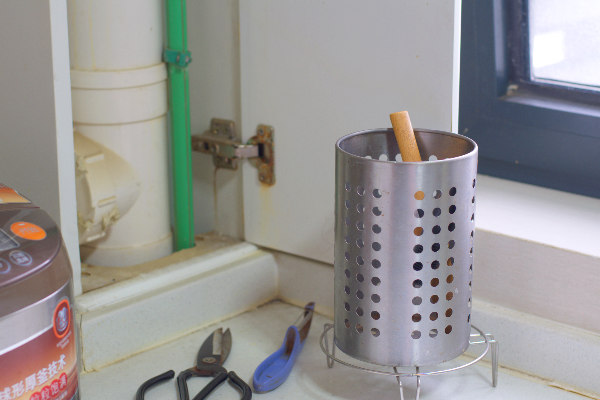}}}
  \centerline{Ground Truth}
\end{minipage}
\begin{minipage}[b]{0.32\linewidth}
  \centering
  \centerline{\adjustbox{trim={0.15\width} {0.1\height} {0.1\width} {0.1\height},clip, width=\textwidth}%
  {\includegraphics[width=\textwidth]{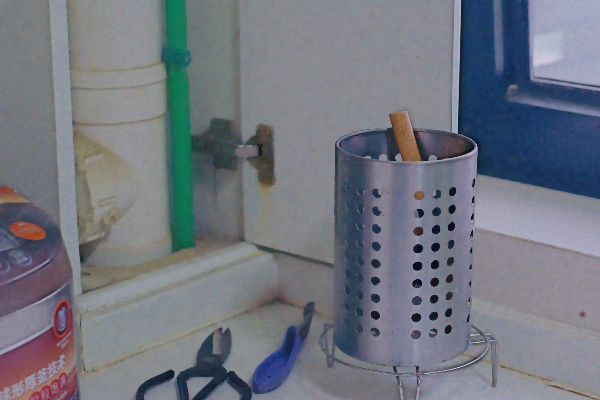}}}
  \centerline{$\text{Tik}(L)+\text{TV}(R)$}
\end{minipage}
\hfill\begin{minipage}[b]{.32\linewidth}
  \centering
  \centerline{\adjustbox{trim={0.15\width} {0.1\height} {0.1\width} {0.1\height},clip, width=\textwidth}%
  {\includegraphics[width=\textwidth]{111Ours.png}}}
  \centerline{$\text{TV}(L)+\text{NLTV}(R)$}
\end{minipage}
\caption{Study on the effect of the priors on the final result. We observe that replacing Tikhonov regularization with total variation for the illumination, combined with a nonlocal total variation prior for the reflectance, leads to improved results in terms of noise reduction and image clarity.}
\label{fig:icip_reg}
\end{figure*}

\section{Ablation Study}

In this section, we conduct an ablation study on the proposed low-light image enhancement method, discussing the influence of the different terms in the variational model \eqref{energia1}, as well as analyzing the pre-processing step. Since most of the novelties proposed in the unfolding counterpart are evaluated using the variational version, we will only study how the designed architecture impacts its performance.

Figure \ref{fig:decomposition} shows the decomposition provided by the proposed method. As expected, the reflectance contains the geometry and texture of the scene with minimal noise, the illumination accurately captures the light intensity, and the noise component does not retain significant structural information.

In Figure \ref{fig:noise_term}, we evaluate the relevance of considering the noise component in the decomposition model. We observe that, when $N$ is omitted, hidden noise in the dark regions is significantly amplified. Instead, our model prevents the enhanced image from being noisy and avoids possible smoothing effects in a post-processing denoising.

Figure \ref{fig:color} illustrates  the impact of the proposed color correction on the low-light image. The adjustment has improved the overall color balance, producing purer white tones. The excessive dominance of the blue channel has been effectively reduced, resulting in a more realistic color distribution.

In Figure \ref{fig:icip_reg}, we observe the effects produced by considering different priors. If we assume a smooth illumination via Tikhonov regularization and TV sparsity on the reflectance, the resulting image is blurred, with noise only partially removed. In contrast, using nonlocal regularization for the reflectance and TV for the illumination, we obtain sharper images with minimal noise. This is one of the key differences compared to the variational model we introduced in our previous conference paper \cite{icip_dani}.

\begin{figure*}[t]
\begin{minipage}[b]{.32\linewidth}
  \centering
  \centerline{\adjustbox{trim={.3\width} {0.4\height} {0.35\width} {0.2\height},clip, width=\textwidth}%
  {\includegraphics[width=\textwidth]{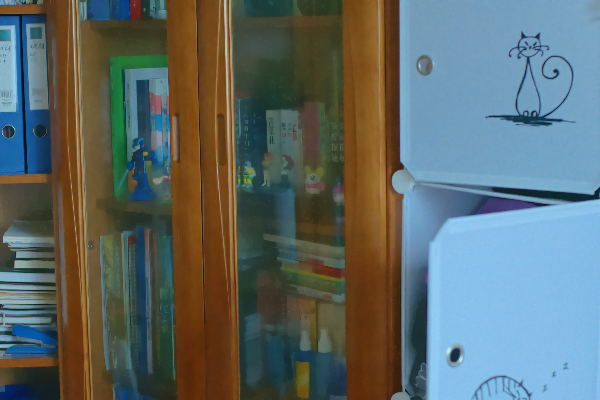}}}
  \vspace{0.1cm}
 \parbox{\textwidth}{\centering No gradient fidelity}
\end{minipage}
\hfill\begin{minipage}[b]{.32\linewidth}
  \centering
  \centerline{\adjustbox{trim={.3\width} {0.4\height} {0.35\width} {0.2\height},clip, width=\textwidth}%
  {\includegraphics[width=\textwidth]{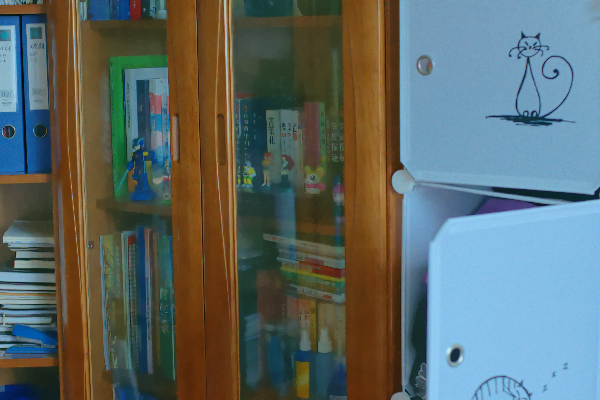}}}
 \vspace{0.1cm}
 \parbox{\textwidth}{\centering $\|\nabla R -\nabla\hat{I}\|_2^2$}
\end{minipage}
\hfill
\begin{minipage}[b]{.32\linewidth}
  \centering
      \vfill
 \centerline{\adjustbox{trim={.3\width} {0.4\height} {0.35\width} {0.2\height},clip, width=\textwidth}%
  {\includegraphics[width=\textwidth]{1Ours.png}}}
 \vspace{0.1cm}
  \parbox{\textwidth}{\centering $\|(\nabla R -\nabla\hat{I})_{\omega}\|_2^2$}
\end{minipage}
\vspace{0.1cm}
\caption{Comparison of the proposed nonlocal gradient-fidelity constraint in \eqref{energia1} with its local variant and the energy model without this term. Although the local version provides some contrast enhancement, it exhibits less defined edges and details compared to our nonlocal approach.}
\label{fig:non_local_term}
\end{figure*}

We also analyze the impact of the newly proposed nonlocal gradient-type constraint in Figure \ref{fig:non_local_term}, comparing its performance with two alternatives: one using a local variant and another without the gradient-fidelity term. We observe that enforcing a gradient constraint enhances edge contrast, but this effect is even more pronounced with our nonlocal approach, resulting in an image with more defined details.

\begin{figure}[t]
\begin{minipage}[b]{0.48\linewidth}
  \centering
  {\includegraphics[width=\textwidth]{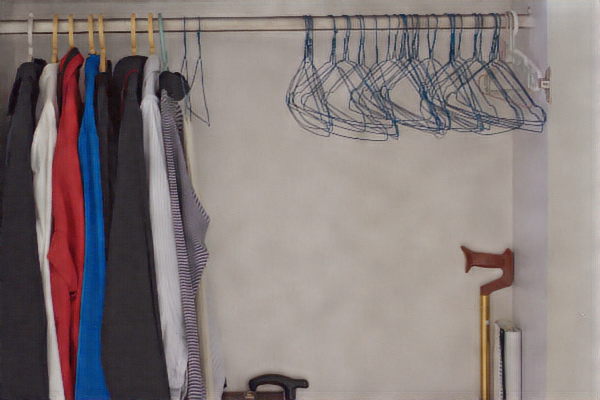}}
 \parbox{\textwidth}{\centering ResNet }
\end{minipage}
\hfill
\begin{minipage}[b]{0.48\linewidth}
  \centering
  {\includegraphics[width=\textwidth]{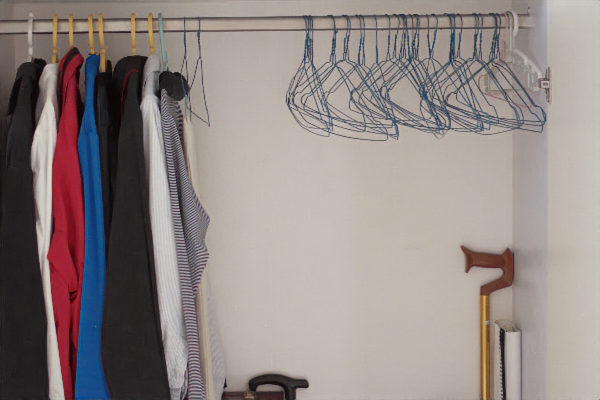}}
  \parbox{\textwidth}{\centering CARNet}
\end{minipage}
\caption{Visual comparison using our unfolding method with CARNet against a modified version with ResNet. The ResNet-based approach exhibits worse color restoration, increased noise, and visible artifacts.}
\label{fig:net-ablation}
\end{figure}

Finally, we assess the contribution of the proposed CARNet to the performance of the unfolding method by replacing it in  \eqref{prox-grad} with a classical fully-convolutional residual network ResNet. The compared ResNet architecture has a larger number of parameters, providing a fair comparison. As seen in Table \ref{table:ablation_resnet} and Figure \ref{fig:net-ablation}, this change significantly deteriorates the results both visually and in terms of all metrics. The final image shows worse color restoration, but the main issue is the persistence of noise and the introduction of artifacts.

\begin{table}[t]
\centering
\begin{tabular}{lccc}
  Architecture   & LPIPS $\downarrow$ & SSIM $\uparrow$ & PSNR $\uparrow$\\ \hline
   ResNet   & 0.163 & 0.8958 & 21.27 \\
   CARNet & 0.108 & 0.9404  & 22.98 \\
\hline
\end{tabular}
\caption{Quantitative evaluation of our unfolding method using CARNet, compared to a modified version with a classical residual network ResNet. The latter shows a significant decrease in performance across all metrics.}
\label{table:ablation_resnet}
\end{table}

\section{Conclusions}

In this work, we have presented a variational method for low-light image enhancement based on the Retinex decomposition of a color-corrected version of the oberserved data into illumination, reflectance, and noise components. Furthermore, the model incorporates a novel nonlocal gradient-based fidelity term, specifically designed to preserve structural details within the image. We also propose integrating our variational formulation into a deep learning framework through an unfolding approach. In this version, the proximal operators are replaced by learnable neural networks. The behaviour of both the nonlocal prior imposed on the reflectance and the nonlocal gradient-type constraint is emulated using cross-attention mechanisms inspired by SWIN transformers.

Our experimental results have shown that the proposed variational method, even without relying on learning-based strategies, performs competitively against state-of-the-art deep learning techniques. Its independence from data-driven training avoids the limitations of requiring paired low-light and ground-truth images. The unfolding approach also achieves superior performance compared to the other techniques, effectively combining the physics-based constraints of variational methods with the learnable priors of deep learning within a flexible and interpretable architecture.

The proposed variational formulation eliminates the need for a post-processing denoising and effectively addresses color degradation. However, its applicability to large-scale datasets is limited by the high computational cost and the large number of parameters involved in the optimization process. On the other hand, the unfolding version is specifically designed for enhancement tasks, but it lacks an explicit mechanism to ensure reliable image decomposition, which should be incorporated in future work.

\bmhead{Acknowledgements}
This work is part of the project TED2021-132644B-I00 funded by MICIU/AEI/10.13039/501100011033 and the European Union NextGeneration EU/PRTR, and also by the MoMaLIP project PID2021-125711OB-I00 funded by MICIU/AEI/10.13039/501100011033 and the European Union NextGeneration EU/PRTR. Daniel Torres is also supported by the Conselleria d’Educació i Universitats del GOIB through grant FPU2024-002-C. The authors gratefully acknowledge the computer resources at Artemisa, funded by the EU ERDF and Comunitat Valenciana and the technical support provided by IFIC (CSIC-UV).

\section*{Declarations}
\subsection*{Data availability}
The first involved dataset LOLv1 \cite{deep-retinex} is publicly available at \url{https://daooshee.github.io/BMVC2018website/}.
The second involved dataset LOLv2-Synthetic \cite{sparse} is publicly available at \url{https://github.com/flyywh/SGM-Low-Light}. The third involved dataset LIME \cite{LIME} is publicly available at \url{https://github.com/aeinrw/LIME/tree/master/data}.

\bibliography{sn-bibliography}

\end{document}